\documentclass[lettersize,journal]{IEEEtran}
\usepackage{amsmath,amsfonts}
\usepackage{algorithmic}
\usepackage{algorithm}
\usepackage{array}
\usepackage[caption=false,font=normalsize,labelfont=sf,textfont=sf]{subfig}
\usepackage{textcomp}
\usepackage{stfloats}
\usepackage{url}
\usepackage{verbatim}
\usepackage{graphicx}
\usepackage{cite}

\usepackage{svg}
\usepackage{multirow}
\usepackage{subfig}
\usepackage{booktabs}

\usepackage{bbding} 
\usepackage{pifont} 
\usepackage{utfsym} 
\usepackage{tabularx} 
\usepackage{hyperref}
\usepackage[table]{xcolor}
\usepackage{colortbl}
\definecolor{lightpink}{rgb}{1.0, 0.9, 0.9}

\usepackage{ulem}

\begin{document}

\title{TBSG-Net: Temporal Bipartite Scene Graph Network for Fine-Grained Video Moment Retrieval}

\author{Ji Huang, Yongsheng Dai, Tianyu Ren, Barry Devereux, Hui Wang, \IEEEmembership{Senior Member, IEEE}
\thanks{This work is supported by the Multimodal Video Search by Examples (MVSE) project funded by UK EPSRC (EP/V002740/2). (\textit{Corresponding author: Hui Wang})}
\thanks{Ji Huang (supported by China Scholarship Council), Yongsheng Dai, Tianyu Ren, and Barry Devereux are with the School of Electronics, Electrical Engineering and Computer Science, Queen's University Belfast, Belfast, Northern Ireland (e-mail: jhuang28@qub.ac.uk; ydai09@qub.ac.uk; tren01@qub.ac.uk; B.Devereux@qub.ac.uk). Ji Huang and Yongsheng Dai contributed equally to this work.}
\thanks{Hui Wang is the director of Discovery AI Lab the NI Landscape Partnership in AI for Bioscience at the School of Electronics, Electrical Engineering and Computer Science, Queen's University Belfast, Belfast, Northern Ireland (e-mail:h.wang@qub.ac.uk)}
}



\maketitle

\begin{abstract}
Recent advances in proposal-free Video Moment Retrieval (VMR) have highlighted the effectiveness of Static Scene Graphs (SSGs). By modeling objects and their relations at the frame level, SSGs enrich retrieval-oriented video representations. However, integrating SSGs into VMR remains constrained by two inherent limitations: (1) Lack of Temporal Dynamics. SSGs fail to model how objects and their relationships evolve over time, leading to the loss of essential temporal dependencies in video representation; and (2) Lack of Explicit Temporal Span Encoding. SSGs do not explicitly encode the duration of relationships, making precise localization challenging. 
To address these limitations, we propose \textbf{\textit{Temporal Bipartite Scene Graph Network} (\textbf{TBSG-Net})}—to the best of our knowledge, the first \textbf{\textit{Dynamic Scene Graph}} \textbf{(DSG)} based proposal-free VMR model. Specifically, TBSG-Net leverages DSGs to extract event-centric graph representations of the input video, enabling the modeling of object interactions over time and thus addressing limitation (1). These DSGs are then processed by a novel \textbf{\textit{Dynamic Scene Graph Embedding} (\textit{DSG-E})} module to capture both Temporal Span and spatio-temporal information. First, DSG-E utilizes a TBSG Constructor to transform DSGs into TBSGs, explicitly encoding objects, relationships, and time spans to tackle limitation (2). Second, the resultant TBSGs are passed into a hybrid TBSG Encoder that integrates a Transformer variant for global event modeling and a Graph Convolutional Network for detailed relational reasoning, ultimately producing a more comprehensive spatio-temporal representation.
\textcolor{black}{Extensive experiments demonstrate substantial improvements 
over all baselines.} \textcolor{black}{TBSG-Net achieves relative gains of 4.30\%, 
11.16\%, and 42.32\% in R@1 at IoU=0.7 on 
Charades-STA, Charades-STA-Len, and 
Charades-STA-Mom respectively, and 
demonstrates cross-dataset generalisability 
via zero-shot transfer to ActivityNet Captions.} TBSG-Net sets a new state-of-the-art, particularly excelling in retrieving
fine-grained moments with intricate temporal dependencies. Code is available: \href{https://github.com/HuangJi1019/TBSG-Net.git}{https://github.com/HuangJi1019/TBSG-Net.git}
\end{abstract}

\section{Introduction}
\begin{figure}[!t]
\centering
\includegraphics[width=3.3in]{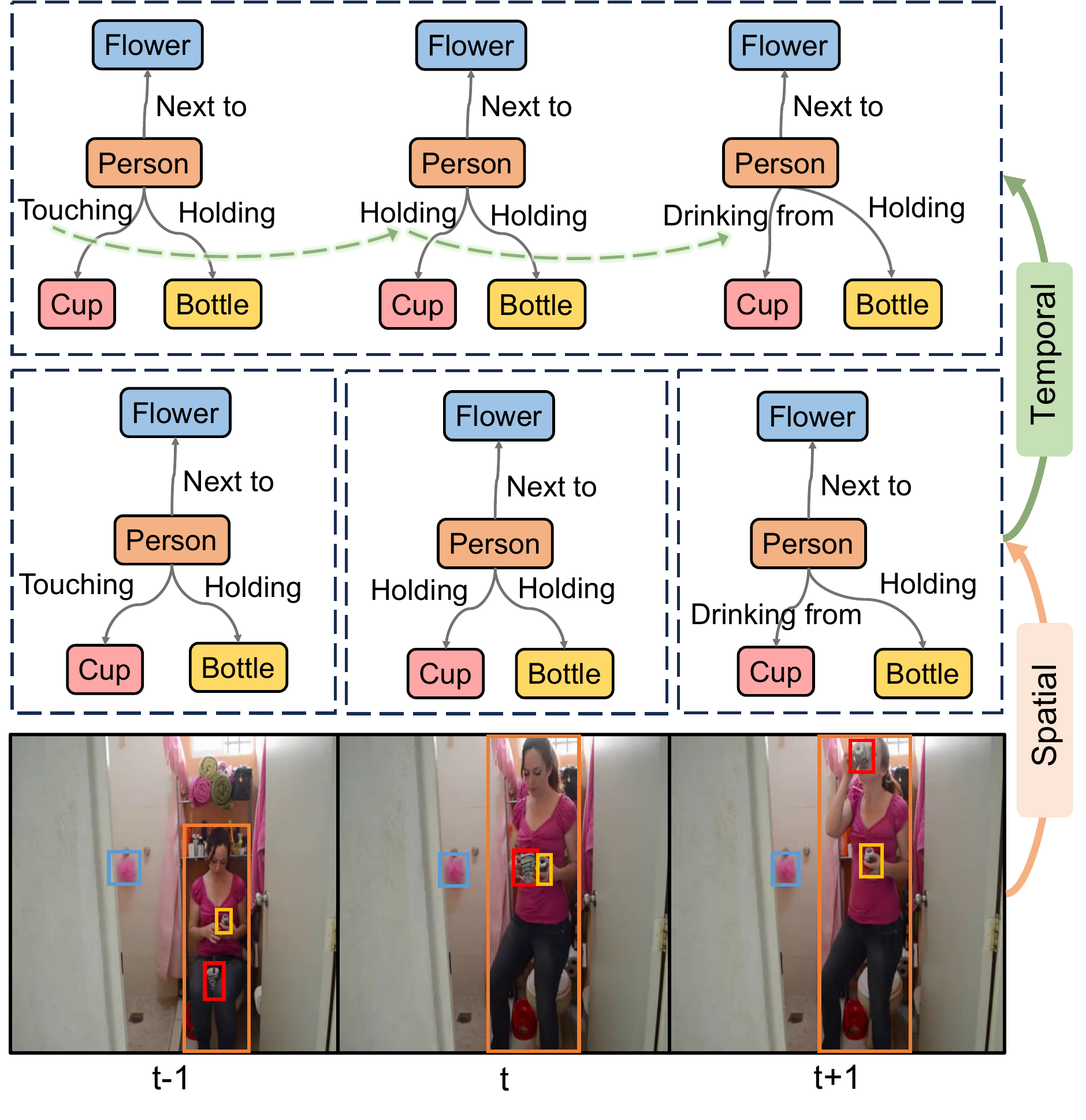}
\caption{DSG Generation Process: (1) The Spatial Module analyzes objects and their spatial relationships across video frames, identifying relationships such as attention, spatial, and contact interactions. (2) The Temporal Module tracks and updates changes in both object and their relationships over time. For clarity, this example highlights the evolution of one relationship with green dashed lines, without depicting changes of the objects and other relationships.
\label{scene graph generation}}
\vspace{-1.5em}
\end{figure}

\IEEEPARstart{V}{ideo} Moment Retrieval (VMR) aims to identify specific temporal segments within an untrimmed video based on a natural language query \cite{gao2017tall}. Unlike static image retrieval, VMR demands understanding of temporal dynamics, scene transitions, and object interactions. This temporal dimension makes VMR applicable 
to various domains, including sketch and shape 
domain~\cite{yuan2023retrieval} and 
entertainment~\cite{yang2024beatdance}. However, it also introduces significant challenges in accurately detecting query-relevant content and precisely localizing the corresponding segments\cite{li2024momentdiff}.

Recent VMR methods adopt the proposal-free paradigm \cite{li2024momentdiff, lei2021detecting}, directly predicting start and end timestamps of target moments. This approach reduces reliance on hand-crafted proposals and improves efficiency compared to traditional proposal-based methods \cite{zhang2020learning,gao2017tall}.
Building on proposal-free VMR, recent work \cite{nguyen2024parallel} introduces Static Scene Graphs (SSGs) to enrich video representations by \textcolor{black}{modeling object-level semantic relationships, 
achieving state-of-the-art performance.}

Although integrating SSGs into proposal-free VMR enhances object-level representations, this approach has two inherent limitations: (1) \textbf{\textit{Lack of Temporal Dynamics.}} Since SSGs are constructed at the frame level, they fail to model how objects and their relationships evolve over time. For example, as shown in Fig. \ref{scene graph generation}, a person holding a cup and later drinking from it constitutes a continuous action. However, SSGs treat these frames as independent images, losing crucial temporal dependencies; (2) \textbf{\textit{Lack of Explicit Temporal Span Encoding}}. SSGs do not explicitly encode the duration of relationships, making it difficult to determine when an interaction begins and how long it lasts. Without these two fundamental capabilities, VMR models struggle to accurately retrieve complex, multi-step events that require fine-grained temporal reasoning. 

To address the aforementioned challenges, we propose the \textbf{Temporal Bipartite Scene Graph Network (TBSG-Net)} (as shown in Fig. \ref{modelgraph}), a novel proposal-free VMR model built on \textbf{Dynamic Scene Graphs} (\textbf{DSGs}) and enhanced with a \textbf{DSG Embedding} (\textbf{DSG-E}) module. TBSG-Net fundamentally differs from SSG-based proposal-free methods by integrating DSGs into VMR. These DSGs capture how objects and their relationships evolve, addressing the challenge of (1) \textit{\textbf{Temporal Dynamics}}. Furthermore, unlike SSGs which process frames independently, DSGs provide a structured representation of event state transitions (e.g., \(\langle person, approaches, cup \rangle \) \(\rightarrow \langle person, drinks \: from, cup \rangle\)), providing the foundation for explicit temporal span encoding.

To explicitly construct the representation 
of the \textit{\textbf{Temporal Span}} and 
spatio-temporal information from DSGs, 
TBSG-Net further processes DSGs through a 
novel DSG-E module, which consists of a TBSG 
Constructor and a hybrid TBSG Encoder. First, 
the TBSG Constructor transforms DSGs into 
TBSGs, which comprise object and relationship 
nodes with temporal span attributes (as shown 
in Fig.~\ref{bipartite graph}). This 
transformation explicitly structures temporal 
span information, laying the foundation for 
fine-grained temporal reasoning. Second, the 
resultant TBSGs are processed by the hybrid 
TBSG Encoder to capture both global and local 
spatio-temporal relationships. The proposed 
encoder integrates a Transformer variant, 
enhanced with a duration-aware temporal 
weighting mechanism and a mask matrix, to 
model global event dependencies. Additionally, 
a Graph Convolutional Network (GCN) aggregates 
local object-object and object-relation 
interactions. This combination ensures global 
event information and fine-grained 
relationships are effectively encoded, thereby 
enhancing retrieval performance for complex, 
multi-step video moments.

In summary, our contributions are as follows:

\begin{itemize}
\item To the best of our knowledge, this is the first work to introduce \textbf{DSGs} into VMR. DSGs provide a graph-structured event representation by modeling objects, their relationships, and the Temporal Dynamics of both, forming the foundation for Temporal Span Encoding.

\item To explicitly encode the Temporal Span and enhance the spatio-temporal representation of DSGs, we propose \textbf{DSG-E}, a novel embedding module comprising the TBSG Constructor and the hybrid TBSG Encoder. The TBSG Constructor converts DSGs into TBSGs, explicitly encoding Temporal Span information. The hybrid TBSG Encoder integrates a Transformer variant for global event modeling and a GCN for fine-grained relational modeling. The resulting embeddings provide a comprehensive spatio-temporal representation, effectively capturing both global event structures and local object interactions.

\item We conduct extensive experiments on 
Charades-STA and its \textcolor{black}{two anti-bias variants, 
and perform zero-shot transfer to ActivityNet 
Captions, demonstrating SOTA performance 
and cross-dataset generalisability.}
\end{itemize}

\section{Related Work}\label{Related Work}
\subsection{Video Moment Retrieval}
\textcolor{black}{VMR is the task of identifying specific 
temporal segments within an untrimmed video that semantically 
align with a given natural language query. Early 
proposal-based approaches generate candidate segments via 
sliding windows~\cite{zhang2020learning}, anchor 
boxes~\cite{gordeev2024saliency}, or learned proposal 
mechanisms, then rank them by semantic 
relevance to the query~\cite{gao2017tall}. However, dense 
candidate sampling introduces substantial computational 
redundancy, particularly for long videos or complex queries.}
To address this, proposal-free methods directly predict the 
start and end timestamps of the target moment. 
\textcolor{black}{MomentDiff~\cite{li2024momentdiff} employs 
a diffusion-based denoiser to iteratively refine temporal 
segments, while UniMD~\cite{zeng2025unimd} unifies moment 
retrieval and temporal action detection within a 
proposal-free encoder--decoder framework.} \textcolor{black}{Beyond 
VMR-specific methods, recent cross-modal retrieval works offer 
complementary insights: ESSE~\cite{wang2024estimating} models 
semantic uncertainty via sector-based geometric representations, 
and SDCMR~\cite{wang2024semantics} disentangles semantic-shared 
from modality-dependent representations for purer cross-modal 
embeddings.}

\begin{figure*}[t]
\centering
\includegraphics[width=5.7in]{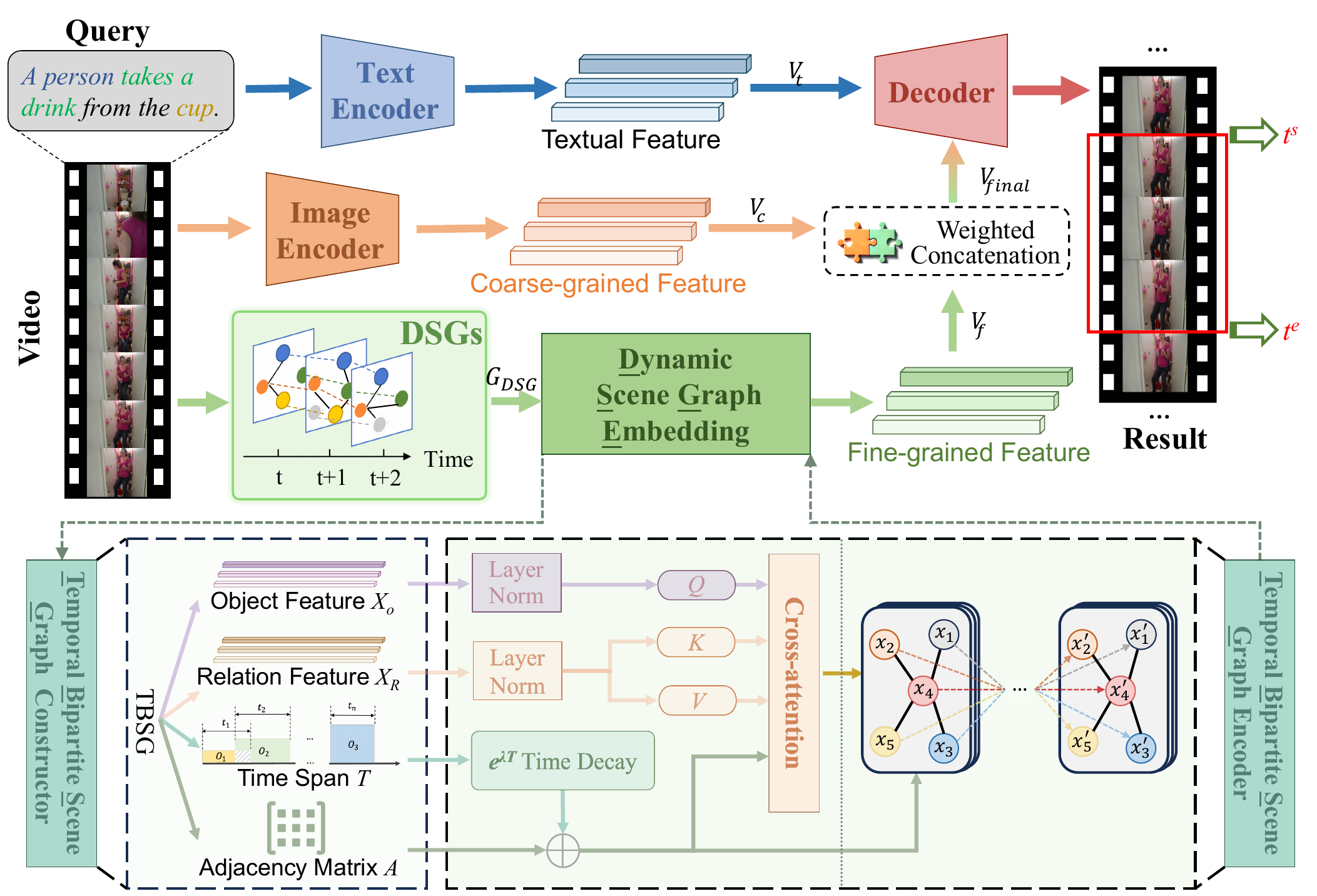}
\caption{The TBSG-Net architecture follows an encoder-decoder paradigm. A Text Encoder first captures the semantic context of the input queries, while the Visual Encoder consists of an Image Encoder for coarse-grained features and Dynamic Scene Graph Embedding (DSG-E) for fine-grained features. The DSG-E module comprises a Temporal Bipartite Scene Graph Constructor (TBSG Constructor) and a hybrid TBSG Encoder. Finally, the Decoder predicts the start and end times of the target video segment, ensuring alignment with the input query.}
\label{modelgraph}
\end{figure*}

\subsection{Scene Graph Generation}
\textcolor{black}{Scene graphs represent objects (nodes) and 
their relationships (edges) as $\langle subject, predicate, 
object \rangle$ triples. Introduced by 
Johnson et al.~\cite{johnson2015image} for image retrieval, 
they have since been applied to image 
captioning~\cite{zhang2022look} and 
retrieval~\cite{yoon2021image}.}

\noindent\textbf{Static Scene Graph.} Static 
scene graphs are designed for single-frame 
analysis, capturing spatial relationships. 
SGTR~\cite{li2022sgtr} introduces a 
transformer-based approach, formulating scene 
graph generation as a bipartite graph 
construction problem, leading to more accurate 
visual relationship modelling. \textcolor{black}{Building on these advances in static SGG, 
QuatRE enriches relation representations by 
embedding relation triplets in quaternion 
space via Hamilton product, 
while CoRE~\cite{wang2022complex} incorporates Hermitian 
inner product to learn relation embeddings in complex space. Both 
demonstrate that expressive mathematical 
spaces significantly improve relation 
modelling in static SGG. Our TBSG-Net extends 
this principle to the temporal domain by 
introducing duration-aware weights 
$\mathbb{T}_{ij}$, bridging the gap between 
static relation embedding and dynamic VMR.}

\noindent \textbf{Dynamic Scene Graph.} DSGs model the evolution of objects and their relationships over time. STTran~\cite{cong2021spatial} 
employs Transformers to capture spatial context within frames 
and decode temporal relationships, modeling object interactions 
over time, \textcolor{black}{and has shown effectiveness in video question 
answering~\cite{cao2023scenegate} and video 
summarization~\cite{zhu2022relational}.} However, their potential in VMR remains largely underexplored. While PaTF \cite{nguyen2024parallel} leverages static scene graphs to enhance VMR, it fails to capture temporal dynamics in videos and does not explicitly encode temporal spans,  leading to suboptimal performance.
To address these limitations, we propose a DSG-based proposal-free VMR model that enables fine-grained event representation and precise temporal localization.

\textcolor{black}{\textbf{Comparison with Graph-based VMR Methods.}
Several prior VMR approaches incorporate graph structures, but 
differ fundamentally from our DSG-based temporal modeling. 
MMRG~\cite{zeng2021multi} builds a proposal-based static 
relational graph that aggregates object and predicate features 
per segment, yielding time-invariant graphs that capture 
neither frame-wise interaction changes nor relational temporal 
spans. MHGR~\cite{wang2024modality} constructs a 
modality-level heterogeneous graph over video, audio, and text 
for multimodal fusion, without explicit subject--predicate--object relational 
structure. In contrast, our DSGs encode object--relation 
evolution over time and are converted into TBSGs with explicit 
relational spans, enabling fine-grained relational--temporal 
reasoning unsupported by static proposal-level or 
modality-level graph formulations.}

\section{Method}\label{Method}
\subsection{Problem Formulation}
Given an untrimmed video \(V = [f_i]_{i=0}^{N_v-1}\) comprising \(N_v\) video frames, and a textual query \(Q = [q_i]_{i=0}^{N_q-1}\) containing \(N_q\) words, the task of VMR aims to localize a specific temporal segment \(x = (t^s, t^e)\) within an untrimmed \(V\) that corresponds to the semantic content of \(Q\). Here, \(t^s\) and \(t^e\) denote the start and end times of the target moment. Formally, the retrieval function can be defined as: 
\begin{equation}
    x = F(Q, V).
\end{equation}
During training, our objective is to minimize the discrepancy between \(x\) and the ground-truth boundaries.

\subsection{Overview}\label{Overview}

\textcolor{black}{As illustrated in Fig.~\ref{modelgraph}, TBSG-Net 
follows a proposal-free encoder-decoder paradigm. The encoder extracts 
representations from both the textual and visual modalities, and the 
decoder predicts the temporal boundaries of the relevant video segment 
via classification and regression heads.}

\textcolor{black}{For text encoding, the input query is encoded into dense semantic 
embeddings using the CLIP text encoder.}
For visual encoding, our framework employs a two-level representation 
strategy: (1) \textbf{Coarse-grained frame-level features:} These 
features capture the global context of individual frames and are 
extracted using pre-trained VGG~\cite{simonyan2014very}, 
CLIP~\cite{radford2021learning}, I3D~\cite{carreira2017quo}, 
\textcolor{black}{or a dual-stream SF+CLIP backbone combining 
SlowFast~\cite{feichtenhofer2019slowfast} motion with CLIP features.} 
(2) \textbf{Fine-grained event-based clip-level representations:} 
These representations focus on representing specific events within 
the video, derived from the DSG Generation process 
(Section~\ref{Dynamic Scene Graph Generation}) and further encoded 
by the DSG Embedding module 
(Section~\ref{Dynamic Scene Graphs Embedding}). 
The two are then combined via weighted concatenation to form the 
final visual representation.

\subsection{Textual and Visual Encoder}\label{Textual and Visual Encoder}
\subsubsection{Text Encoder}
The text encoder \( E_{\text{t}} \), derived from CLIP, encodes textual queries into dense semantic embeddings. Given a query \( Q = [q_i]_{i=0}^{N_q-1} \), \( E_{\text{t}} \) produces token-level embeddings in \(\mathbb{R}^{N_q \times D}\). Additionally, a sentence-level feature \( V_t \in \mathbb{R}^D \) is extracted from the End of Sentence ([EOS]) token, summarizing the holistic semantic content of the query.

\subsubsection{Video Encoder}\label{Video Encoder}
\textcolor{black}{Our video encoder extracts the two levels of visual representations 
described in 
Section~\ref{Overview}.} \textcolor{black}{Specifically, a pre-trained 
backbone $F_b$ is employed to extract coarse-grained features from 
the raw video sequence $V = [f_i]_{i=0}^{N_v-1}$:}
{\small
\begin{equation}
    \textcolor{black}{V_b = F_b([f_i]_{i=0}^{N_v-1}), \quad
    b \in \{\text{VGG}, \text{CLIP}, \text{I3D}, 
    \text{SF+CLIP}}\},
\end{equation}
}
\noindent where $V_{\text{VGG}}$ captures static appearance features 
from individual video frames, $V_{\text{CLIP}}$ obtains semantically 
rich representations that are inherently aligned with textual 
semantics, $V_{\text{I3D}}$ models short-term motion dynamics in 
videos, \textcolor{black}{and $V_{\text{SF+CLIP}}$ combines SlowFast 
motion features with CLIP features for richer temporal 
representation.}

In parallel to these coarse-grained features, we construct dynamic 
scene graphs $G_{\text{DSG}}$ for fine-grained clip-level 
representations. This DSG generation module comprehensively models 
the evolution of objects and their relationships over time 
(Section~\ref{Dynamic Scene Graph Generation}). The resulting dynamic 
scene graphs are then encoded by the Dynamic Scene Graph Embedding 
(DSG-E) module (Section~\ref{Dynamic Scene Graphs Embedding}) to 
produce the fine-grained spatio-temporal representation:
{\small
\begin{equation}
\label{TBSGE}
V_f = F_{\text{DSG-E}}(G_{\text{DSG}}(V)) = F_{\text{DSG-E}}(G_{\text{DSG}}([f_i]_{i=0}^{N_v-1}))\color{black}.\color{black}
\end{equation}
}
The final video representation $V_{\text{final}}$ is obtained through 
a weighted concatenation of the fine-grained feature $V_f$ and the 
chosen coarse-grained feature 
$\textcolor{black}{V_g} \in \{V_{\text{VGG}}, V_{\text{I3D}}, 
V_{\text{CLIP}}, \textcolor{black}{V_{\text{SF+CLIP}}}\}$:
{\small
\begin{equation}
\label{final_emb}
V_{\text{final}} = [W_f \cdot V_f \;;\; \textcolor{black}{W_g \cdot V_g}]\color{black},\color{black}
\end{equation}
}
\noindent where \([;]\) denotes the concatenation operation along the feature dimension, and \(W_f\) and \textcolor{black}{$W_g$} are learnable parameters. 

\subsection{Dynamic Scene Graph Generation}\label{Dynamic Scene Graph Generation}

Given an untrimmed video \( V = [f_i]_{i=0}^{N_v-1} \) comprising \(N_v\) frames, we construct DSGs \(G_{\text{DSG}}\) using the STTran \cite{cong2021spatial} model, as illustrated in
Fig.~\ref{scene graph generation}. STTran is designed to model the evolution of objects and their relationships over time \textcolor{black}{ and serves as an off-the-shelf
relational detector for dynamic scene graph generation}.

\textcolor{black}{Although STTran is adopted in our implementation as an upstream relational
detector, our framework does not rely on dataset-specific heuristics or
handcrafted priors. When transferring TBSG-Net to a new dataset, the same
pipeline can be applied by training the relational detector using the dataset’s
available object and relation annotations during offline preprocessing, without
modifying the overall architecture.} The STTran model contains two main stages: 

(1) The spatial module is built upon a Faster R-CNN object detector. Specifically, it detects object features \(V_{v,t} = \{v_t^0, \ldots, v_t^{N(t)-1}\}\), bounding boxes \(V_{b,t} = \{b_t^0, \ldots, b_t^{N(t)-1}\}\), and object classes \(V_{c,t} = \{c_t^0, \ldots, c_t^{N(t)-1}\}\), where \(N(t)\) represents the number of detected objects in frame \(t\). A spatial transformer with positional encoding analyzes intra-frame object
relationships and outputs both relational class labels and relation feature
embeddings,
denoted as $R_{r,t} = \{r_t^0, \ldots, r_t^{K(t)-1}\}$ and
$R_{v,t} = \{u_t^0, \ldots, u_t^{K(t)-1}\}$ respectively,
where $K(t)$ denotes the number of identified relationships in frame $t$.

(2) The temporal module then operates on the output of the spatial module, modeling how detected objects and their relationships evolve across frames. By capturing the temporal dependencies within objects and their relationships, it ensures that \(G_{\text{DSG}} = (V_{c,t}, R_{r,t})\)  effectively encodes spatial-temporal relationships present in the video. The corresponding visual features 
\((V_{v,t}, V_{b,t}, R_{v,t})\) are retained alongside the graph structure for 
subsequent node feature construction in the TBSG (Section \ref{Dynamic Scene Graphs Embedding}). \textcolor{black}{The resulting $G_{\text{DSG}}$ is then passed 
to the DSG Embedding module for temporal 
bipartite scene graph construction and encoding.}

\subsection{Dynamic Scene Graph Embedding}\label{Dynamic Scene Graphs Embedding}
\subsubsection{Temporal Bipartite Scene Graph Constructor}\label{Temporal Bipartite Scene Graph Constructor}
\textcolor{black}{The TBSG represents a video 
clip as a bipartite graph with two disjoint 
node sets: object nodes $O$ and relation nodes 
$R$. Edges exist exclusively between $O$ and 
$R$, encoding the subject-predicate-object 
structure of each detected interaction. This 
bipartite formulation enables the TBSG Encoder 
to reason separately over object-level 
appearance and relation-level semantics before 
producing a unified clip representation.}

\begin{figure}[!t]
\centering
\includegraphics[width=3in]{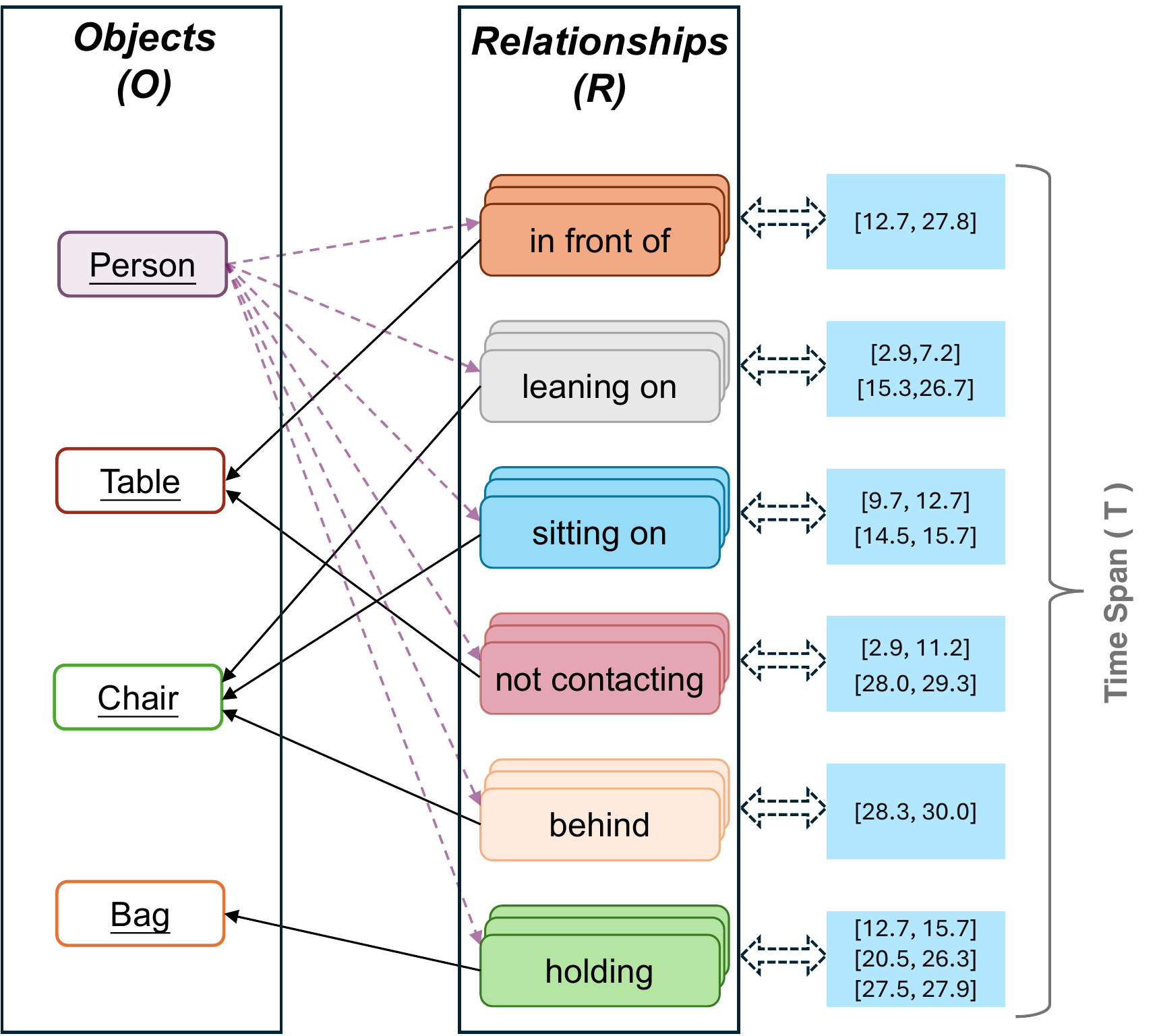}
\caption{Demonstration of TBSGs. Nodes are divided into objects and relationships, with each relationship node having a time span attribute (in seconds). Dashed lines denote subjects to relationships, solid lines connect relationships to objects. Each time span marks the duration of every triplet \(\langle subject, relationship, object \rangle\).} 
\label{bipartite graph}
\end{figure}

To explicitly model the temporal spans of 
objects and their relationships, we transform 
the Dynamic Scene Graph $G_{\text{DSG}} = 
(V_{c,T'}, R_{r,T'})$ along with its associated 
visual features $(V_{v,T'}, V_{b,T'}, R_{v,T'})$ 
into a Temporal Bipartite Scene Graph for a 
video clip $C_{T'}$ spanning $T'$ frames, as 
illustrated in Fig.~\ref{bipartite graph}. 
First, we compute the union sets of unique 
objects and relationships that appear within 
these $T'$ frames. Let $V_{c,1:T'} = 
\bigcup_{t=1}^{T'} V_{c,t}$, $R_{r,1:T'} = 
\bigcup_{t=1}^{T'} R_{r,t}$; here, 
$V_{c,1:T'}$ (denoted as set $O$) contains 
all unique objects, and $R_{r,1:T'}$ (denoted 
as set $R$) contains all unique relationships 
observed in the clip. \textcolor{black}{In the Charades-STA benchmark and its variants}, 
the entity \textit{Person} is the only subject 
class, and we treat it as an object in 
subsequent discussions.

To model how interactions evolve over time, 
we connect objects and relationships across 
frames and associate each relationship with a 
timestamp. Formally, the $e$-th connection 
instance is defined as: 
\textcolor{black}{\small
\begin{equation}
\label{edge}
    E^{(e)} = (o^{(e)},\, r^{(e)},\, t^{(e)}), 
\quad e = 1, \ldots, |\mathcal{E}|,
\quad o^{(e)} \in O,\; r^{(e)} \in R
\color{black},\color{black}
\end{equation}
}
\noindent where $t^{(e)}$ denotes the timestamp 
of the $e$-th connection instance. 
\textcolor{black}{Here $e$ indexes the edge 
instances rather than nodes within $O$ or $R$; 
the object and relation node indices are written 
as $i \in \{1,\ldots,|O|\}$ and $j \in 
\{1,\ldots,|R|\}$ respectively, where $|O|$ 
and $|R|$ are independently determined by the 
clip content and are in general unequal. The 
set $\{E^{(e)}\}$ serves as the common source 
for both the temporal structure and the graph 
topology of the TBSG: the timestamps $t^{(e)}$ 
are aggregated into temporal spans as described in the remainder 
of this section, while the index pairs 
$(o^{(e)}, r^{(e)})$ determine the connectivity 
structure of $A_{or}$ (Eq.~\ref{TBG}). We define the set of all temporal spans within 
a clip as:} 
{\small
\begin{equation}
\label{timestamps} 
\mathcal{T} = \{[t_{\text{s}}, t_{\text{e}}] \mid 0 \leq t_{\text{s}}, t_{\text{e}} \leq T_{\text{max}}, t_{\text{s}} < t_{\text{e}}\}\color{black},\color{black}
\end{equation}
}
\noindent \textcolor{black}{where $T_{\max}$ denotes the 
dataset-wide maximum video duration. If a 
relationship persists across consecutive 
frames, its timestamps are merged into a 
single continuous span $[t_s, t_e]$, which is 
recorded as one entry in $\mathcal{T}$, 
representing the interaction more compactly 
for event-level analysis. For each 
object-relation pair $(i,j)$, the scalar 
temporal span $\mathcal{T}_{ij}$ is derived 
by aggregating the relevant entries of 
$\mathcal{T}$. Specifically, for cases where 
the same pair appears in multiple disjoint 
intervals within a clip, each contiguous 
segment is first recorded as an independent 
span in $\mathcal{T}$, and the final 
$\mathcal{T}_{ij}$ is computed as 
$\mathcal{T}_{ij} = \sum_{k}(t_e^k - t_s^k)$, 
where $[t_s^k, t_e^k]$ denotes the $k$-th 
contiguous interval in which the pair is 
active within the clip.}

We denote the node feature of the TBSG as \(X_o\) for objects in \(O\) and \(X_r\) for relationships in \(R\). \textcolor{black}{The initial object node feature \(X_o\)
combines object category, visual feature, and spatio-temporal localization
information. Specifically, \(V_{c,t}\), \(V_{v,t}\), and \(V_{b,t}\) denote the
object class label, visual feature, and bounding box feature at time step \(t\),
respectively, all extracted by the STTran 
model~\cite{cong2021spatial}. \textcolor{black}{Here $\phi(\cdot)$ denotes the CLIP text encoder applied to 
categorical label names (e.g., ``cup'', ``person''), 
converting discrete class labels into continuous semantic 
embeddings that are aligned with the visual feature space, $\varphi(\cdot)$ denotes a linear 
projection, and $V_u$ is the union box feature 
computed via RoIAlign.} For notational simplicity, we omit the time subscript $t$ when 
the context is clear.} The object node feature is defined as:
{\small
\begin{equation}
    X_o = \langle W_c\phi(V_{c}), W_vV_v, W_b \varphi( V_u \oplus F_{\text{box}}(V_{b})) \rangle \color{black},\color{black}
\end{equation}
}\textcolor{black}{where $\langle \cdot, \cdot, \cdot \rangle$  denotes concatenation followed by a linear
projection,} \(\oplus\) represents element-wise addition, \( F_{\text{box}} \) transforms bounding boxes into feature representations, \textcolor{black}{and $W_c$, $W_v$, $W_b$ are learnable 
projection matrices}. Similarly, the relationship node feature is defined as:
{\small
\begin{equation}
    X_r = \langle W_r\phi(R_r), W_r'R_v \rangle \color{black},\color{black}
\end{equation}
}
\noindent where $R_r$ denotes the relation 
type label (e.g., ``holding''), $R_v$ 
denotes the corresponding visual relation 
feature extracted by the 
STTran~\cite{cong2021spatial}, \textcolor{black}{and \(W_r\), \(W_r'\) are learnable projection 
matrices.}  \textcolor{black}{Specifically, 
$R_v$ is computed using distinct projection 
matrices $\mathbf{W}_s$ and $\mathbf{W}_o$ 
applied respectively to the subject and object 
RoI features (Eq.~3 in~\cite{cong2021spatial}), 
such that $X_r$ inherently encodes asymmetric 
role information for each detected interaction.}
  
Finally, we define the TBSG for the clip as:
\textcolor{black}{{\small
\begin{equation}
\label{TBG}
   G_C = \{X_o, X_r, \mathcal{T}_{ij}, A_{or}\},
\end{equation}
}}
\noindent where $A_{or}$ is the adjacency 
matrix that records valid object-relationship 
connections. Specifically, $A_{or}[i,j] = 1$ 
if object $i$ participates in relation $j$ 
(as either subject or object) in any frame 
of the clip, and $A_{or}[i,j] = 0$ otherwise. 
\textcolor{black}{Here $\mathcal{T}_{ij}$ 
denotes the scalar temporal span for the 
specific pair $(i,j)$, computed as $\mathcal{T}_{ij} = 
\sum_k(t_e^k - t_s^k)$ above. Note 
that $A_{or}$ encodes only structural 
connectivity; semantic directionality is 
preserved separately in $X_r$ as described 
above.}

\textcolor{black}{In practice, transient detection errors are 
naturally suppressed during temporal aggregation, as interactions 
persisting across multiple frames are preferentially retained over 
isolated one-frame predictions.}

\subsubsection{Temporal Bipartite Scene Graph Encoder}\label{Temporal Bipartite Scene Graph Encoder} 

\textcolor{white}{}\\
\noindent\textbf{Transformer variant module.}\label{Transformer variant module.} 
To capture temporal dynamics within the bipartite 
graph, we introduce a duration-aware temporal 
weighting term
\textcolor{black}{$\mathbb{T}_{ij} = \exp\!\left(\lambda \cdot 
\frac{\mathcal{T}_{ij}}{T_{\max}}\right)$,
where $\mathcal{T}_{ij}$ is the scalar temporal 
span for pair $(i,j)$ defined in 
Section~\ref{Temporal Bipartite Scene Graph Constructor}, 
$T_{\max}$ is the maximum video duration 
(Eq.~\ref{timestamps}), and $\lambda$ 
is a learnable scalar parameter initialised at 
$0.1$ and optimised end-to-end, controlling the 
sensitivity to interaction duration.} This term 
ensures that the model places greater emphasis 
on longer interactions while progressively 
reducing the impact of shorter relationships.

We formulate the attention in our Transformer as follows:
{\small
\begin{equation}
      \text{Attention}_{O \rightarrow R} = \text{softmax}\left(\frac{H_o H_r^T}{\sqrt{d_k}} + M +  \mathbb{T}\right)H_r\color{black},\color{black}
\end{equation}
}
\noindent 
where \(d_k\) represents the dimension of the key, \(H_o\) and \(H_r\) are the transformed feature representations of object set \(O\) and relationship set \(R\) respectively, \textcolor{black}{and $\mathbb{T} \in \mathbb{R}^{|O| \times |R|}$ is the 
duration-aware weight matrix with entries $\mathbb{T}_{ij}$ 
defined above.} \textcolor{black}{$H_o$ and $H_r$ are computed as}:
{\small
\begin{equation}
\begin{gathered}
    H_o = X_o \hat{W}_o, \quad H_r = X_r \hat{W}_r,
\end{gathered}
\end{equation}
}
\noindent where \(\hat{W}_o\) and \(\hat{W}_r\) are learnable matrices that project the features \(X_o\) and \(X_r\) into a common embedding space.

To ensure that attention is computed only for existing edges in the bipartite graph, a mask matrix \(M\) is applied:
{\small
\begin{equation}
    M_{ij} = 
\begin{cases} 
0 & \text{if } A_{or}[i, j] = 1 \\
-\infty & \text{otherwise}
\end{cases} \color{black}.\color{black}
\end{equation}
}
 This attention mask strictly enforces selective attention, allowing the Transformer to focus solely on valid existing connections in the bipartite graph. \textcolor{black}{To stabilise training, a residual connection 
and layer normalisation are applied to the output of the 
attention module before passing features to the GCN. The resulting features are denoted as 
$H_{\text{Attn}}$ and serve as the input to the subsequent GCN module.}

\noindent\textbf{Graph Convolutional Networks Module.}
To capture local neighborhood structures, we apply GCN, which iteratively aggregates and propagates features across connected nodes.  
\textcolor{black}{Specifically, 
each edge weight $A_{or}[i,j]$ is scaled by 
$\mathbb{T}_{ij}$ to form $A'_{or}[i,j] = 
A_{or}[i,j] \cdot \mathbb{T}_{ij}$, and 
$\tilde{A}'_{or}$ is its symmetrically 
normalised form, with degree matrix 
$\tilde{D}_{ii} = \sum_j \tilde{A}'_{or,ij}$.} The GCN update at layer \(l\) then becomes:
 
{\small
\begin{equation}
    H^{(l+1)} = \sigma\left(\tilde{D}^{-\frac{1}{2}} \tilde{A}'_{or} \tilde{D}^{-\frac{1}{2}} H^{(l)} W^{(l)}\right) \color{black},\color{black}
\end{equation}
}
\noindent \textcolor{black}{$H^{(l)}$ is the node feature matrix 
at GCN layer $l$, initialised as $H^{(0)}=H_{\text{Attn}}$, 
where $H_{\text{Attn}}$ denotes the output of the preceding 
Transformer submodule within the TBSG Encoder, after residual 
connection and layer normalisation. Each layer has a learnable 
weight $W^{(l)}$ and a nonlinear activation $\sigma$.}

\begin{figure}[t]
\centering
\subfloat[]{\includegraphics[width=0.1\textwidth]{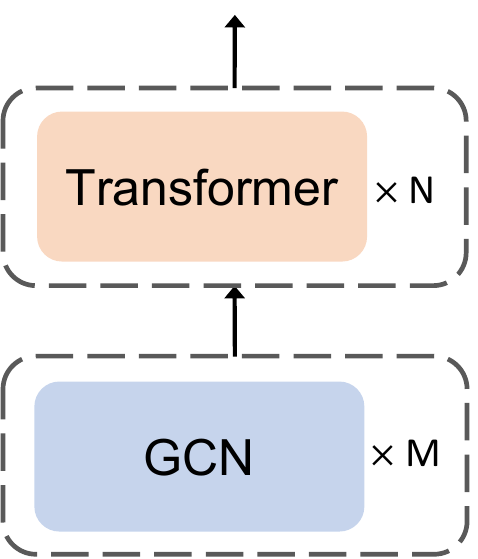}\label{fig:sub1}}\hfill
\subfloat[]{\includegraphics[width=0.1\textwidth]{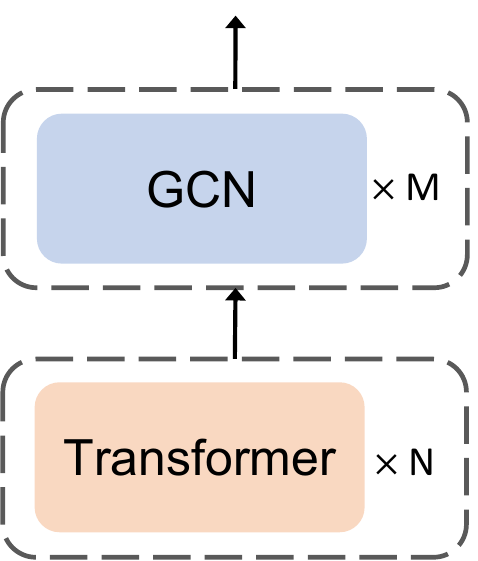}\label{fig:sub2}}\hfill
\subfloat[]{\includegraphics[width=0.2\textwidth]{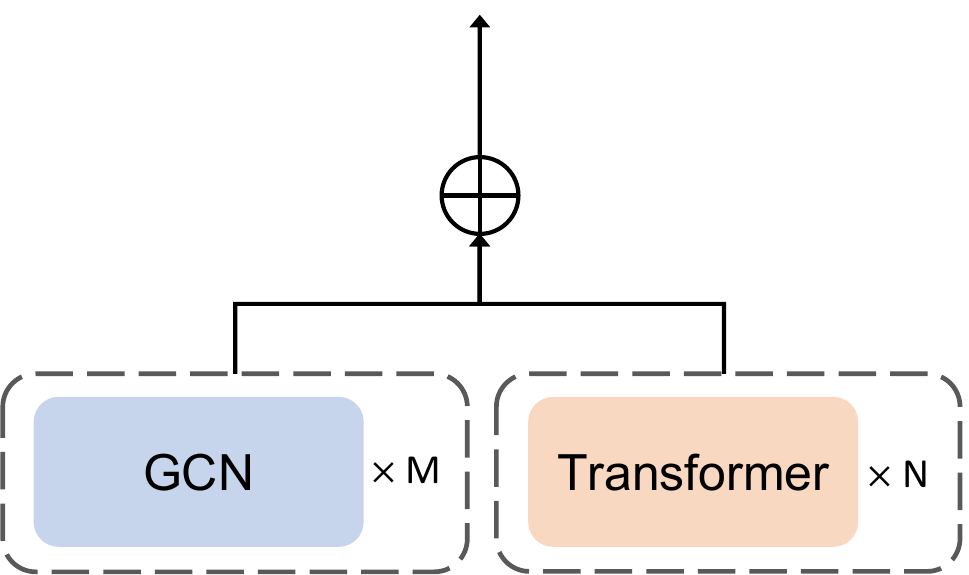}\label{fig:sub3}}

\caption{Three configurations of relative positioning in the Transformer-GCN.}
\label{fig:relative_position}
\end{figure}

\noindent\textbf{Architecture variants.}
We explore three types of relative position of Transformer-GCN architecture (Fig. \ref{fig:relative_position}). These variants differ in how features are passed and combined between Transformer-based attention and GCN-based neighborhood aggregation. Quantitative comparisons, shown in Table \ref{tab:relative position of Tran-GCN}, confirm that placing these modules in different orders can significantly affect performance. 

\textcolor{black}{The TBSG Encoder improves robustness by 
leveraging relational and global contextual reasoning. GCN propagation 
allows nodes with noisy features to be refined by structurally consistent 
neighbors, while the Transformer integrates long-range relational–temporal 
dependencies that reduce the influence of isolated errors. Together, TBSG 
Constructor and TBSG Encoder reasoning ensure that the final representation 
focuses on stable relational–temporal structure rather than frame-level 
noise. Importantly, this robustness is achieved through general architectural 
design rather than dataset-specific tuning.}

\textcolor{black}{The TBSG Constructor and TBSG Encoder operate based on 
graph topology and temporal structure, without relying on dataset-specific 
heuristics or handcrafted semantic priors. This design enables direct 
application to any dataset once object–relation or scene-graph inputs 
become available.}

\subsubsection{Training Objectives}\label{Training Objectives}
To effectively train TBSG-Net for video moment retrieval, we adopt a composite loss function that guides the learning of both temporal relevance and precise boundary prediction. Following the approach of UniMD \cite{zeng2025unimd}, our training objective comprises a classification loss $L_{cls}$ and a regression loss $L_{reg}$, which are applied to each predicted temporal segment. The overall loss is formulated as follows:
{\small
\begin{equation}
L = \sum_t \sum_i \Bigl(\lambda_{cls}\,L_{cls}(p_{t,i}, y_{t,i})
  + \delta_{t,i}\,\lambda_{reg}\,L_{reg}(b_{t,i}, g_{t,i})\Bigr) \color{black},\color{black}
\end{equation}
}
\noindent where \(t\) indexes the predicted temporal segments, \(i\) iterates over the different scales of these predictions, $\lambda_{cls}$ and $\lambda_{reg}$ are weighting coefficients that balance the relative importance of the classification and regression objectives, and $\delta_{t,i}$ is a binary indicator function such that $\delta_{t,i} = 1$ if the \(i\)-th prediction at temporal prediction point \(t\) is considered a positive sample (i.e., it overlaps sufficiently with a ground-truth moment), and \(\delta_{t,i} = 0\) otherwise.

\noindent\textbf{Classification Loss $L_{cls}$:} We employ a binary classification loss to predict the semantic relevance of each temporal segment to the given textual query. Specifically, for each segment $t$ and scale \(i\), $p_{t,i}$ represents the predicted probability of the segment being relevant to the query, and $y_{t,i} \in \{0, 1\}$ is the corresponding ground-truth label (1 for relevant, 0 for irrelevant). The binary cross-entropy loss is used here:
{\small
\begin{equation}
L_{cls}(p, y) = - (y \log(p) + (1 - y) \log(1 - p))\color{black}.\color{black}\end{equation}
}
\noindent\textbf{Regression Loss $L_{reg}$:} To refine the 
temporal boundaries of the relevant moments, we use a 
regression loss that measures the discrepancy between the 
predicted boundaries $b_{t,i} = (t^{\text{pred}}_s, 
t^{\text{pred}}_e)$ and the ground-truth boundaries 
$g_{t,i} = (t^{\text{gt}}_s, t^{\text{gt}}_e)$. Following 
UniMD and common practices in temporal localization, we employ 
IoU loss: 
{\small
\begin{equation}
L_{reg}(b, g) = 1 - \text{IoU}(b, g),\end{equation}
}
where 
$\text{IoU}(b, g) = {\text{length}(b \cap g)}/{\text{length}
(b \cup g)}$.

\section{Experiments}\label{Experiments}

\subsection{Datasets and Evaluation Metrics}

\noindent\textbf{Charades-STA Dataset.} The Charades-STA dataset (left part of Table \ref{detailed_information_dataset}) extends the Charades dataset \cite{gao2017tall} by providing temporal annotations for queries. Uniquely, Charades-STA includes rich scene graph annotations across 234,253 frames, covering 35 distinct object classes and 25 relationship categories, which are further grouped into three types: attention, spatial, and contact relationships. We adopt Charades-STA as our primary benchmark because it is currently the only
widely used VMR dataset that supports reliable construction of scene-graph--based
relational inputs.
Other popular benchmarks such as QVHighlights and TACoS do not provide compatible object--relation annotations, which prevents a fair and meaningful evaluation of
DSG-based relational modeling. \textcolor{black}{Although ActivityNet Captions likewise lacks native 
object--relation annotations, we construct pseudo-label DSGs 
using the pre-trained STTran~\cite{cong2021spatial} model for zero-shot transfer 
evaluation (Section~\ref{Zero-Shot 
Generalization to ActivityNet Captions}).} This evaluation protocol is consistent with prior scene-graph--based VMR work~\cite{nguyen2024parallel}, which also reports results on
a single dataset due to the limited availability of benchmarks supporting
structured relational inputs.

{\small
\begin{table*}
\centering
\renewcommand\arraystretch{1} 
\caption{Detailed information of the Charades-STA, Charades-STA-Len, and Charades-STA-Mom datasets. Here, $w_0$ is the duration of the target moment and $c_0$ is its center time.
The expressions $c_0 + \tfrac{w_0}{2}$ and $c_0 - \tfrac{w_0}{2}$ denote the end time and start time, respectively.
\label{detailed_information_dataset}}

\begin{tabular}{
  >{\centering\arraybackslash}p{1.4cm}|
  >{\centering\arraybackslash}p{1.8cm}|
  >{\centering\arraybackslash}p{1.4cm}
  >{\centering\arraybackslash}p{1.4cm}|
  >{\centering\arraybackslash}p{1cm}|
  >{\centering\arraybackslash}p{2.2cm}
  >{\centering\arraybackslash}p{2.2cm}|
  >{\centering\arraybackslash}p{1cm}
}
\toprule
\multirow{2}{*}{Dataset} & Charades-STA &\multicolumn{3}{c|}{Charades-STA-Len} & \multicolumn{3}{c}{Charades-STA-Mom}  \\
\cline{2-8}
&Total &$w_0 \leq 10s$ & $w_0 > 10s$ & Total & $c_0 +\frac{w_0}{2} \leq 15s$ & $c_0 - \frac{w_0}{2} > 15s$ & Total \\
\midrule
Training  &12408     & 9307   & 2326  & 11633  & 5330  & 1332  & 6662\\
Test  &3720   & 197     & 788     & 985   & 259    & 1038   & 1297 \\
\bottomrule
\end{tabular}
\end{table*}
}

\noindent\textbf{Charades-STA-Len Dataset.} As introduced in \cite{li2024momentdiff}, the Charades-STA-Len variant modifies the moment duration distribution of the target moments within the Charades-STA dataset. As shown in the middle part of Table \ref{detailed_information_dataset}, the training split is biased toward shorter temporal moments (duration $w_0\leq$ 10 seconds), while the test split predominantly contains longer moments (duration $w_0>$ 10 seconds). This controlled distribution shift enables a rigorous evaluation of the model's ability to generalize retrieval performance across varying moment duration.

\noindent\textbf{Charades-STA-Mom Dataset.} Also proposed in \cite{li2024momentdiff}, the Charades-STA-Mom variant targets the location of moments within Charades-STA. Moments are split at their center time ($c_0$), as shown in the right part of Table~\ref{detailed_information_dataset}, resulting in a training set of earlier moments (before 15 seconds) and a test set of later moments (after 15 seconds). This design creates a particularly challenging scenario for temporal generalization, requiring models to accurately and robustly locate moments regardless of their position within the video timeline.

\noindent\textbf{Evaluation Metrics.} We assess retrieval performance using R@$k$ at IoU thresholds of $\mu$. This metric measures the fraction of ground-truth moments correctly retrieved within the top-$k$ predictions (IoU$\ge\mu$). We report results for $k \in \{1, 5\}$ and IoU thresholds $\mu\in\{0.3$, $0.5$, $0.7\}$. 
{\small
\begin{table*}[!t]
\caption{Performance comparison between 
TBSG-Net and SOTA models on Charades-STA. 
Results are reported for R@$k$, IoU=$\mu$ 
($k \in \{1, 5\}, \mu\in\{0.3, 0.5, 0.7\}$), 
\textcolor{black}{grouped by visual backbone: VGG, CLIP, I3D, 
and SF+CLIP, where SF+CLIP denotes a 
dual-stream backbone combining SlowFast and 
CLIP features. The best and second-best 
results are highlighted in \textbf{bold} and 
\underline{underlined}, respectively. 
The arrows $\uparrow$/$\downarrow$ indicate 
the relative improvement/gap compared to the 
best external baseline in each column. For 
the VGG group, both TBSG-Net variants 
(GloVe and CLIP text encoders) are reported 
to demonstrate the contribution of the text 
encoder.}}
\label{tab:main-result}
\centering
\renewcommand\arraystretch{1.1} 
\begin{tabular}{
 >{\centering\arraybackslash}p{2.3cm}| 
 >{\centering\arraybackslash}p{1.8cm}| 
 >{\centering\arraybackslash}p{1.3cm}| 
 >{\centering\arraybackslash}p{1.5cm}| 
 >{\centering\arraybackslash}p{0.9cm}  
 >{\centering\arraybackslash}p{0.9cm}  
 >{\centering\arraybackslash}p{0.9cm}| 
 >{\centering\arraybackslash}p{0.9cm}  
 >{\centering\arraybackslash}p{0.9cm}  
 >{\centering\arraybackslash}p{0.9cm} 
}

\toprule
\multirow{2}{*}{Method} &\multirow{2}{*}{Venue}
&\multirow{2}{*}{Backbone} &\multirow{2}{*}{Text Encoder}
&\multicolumn{3}{c|}{R@1, IoU=$\mu$} &\multicolumn{3}{c}{R@5, IoU=$\mu$}\\ 

\cline{5-10}

& & & &$\mu$ = 0.3 &$\mu$ = 0.5 &$\mu$ = 0.7 &$\mu$ = 0.3 &$\mu$ = 0.5 &$\mu$ = 0.7\\

\midrule

DEBUG \cite{lu2019debug} &EMNLP-IJCNLP 2019 &VGG &Glove &54.95 &37.39 &17.69 & - &- &- \\
MAN \cite{zhang2019man} &CVPR 2019 &VGG &Glove &- &41.24 &20.54 & - &83.21 &51.85\\
DORi \cite{rodriguez2021dori} &WACV 2021 & VGG &Glove &\underline{61.83} &43.47&26.37 & - &- &-\\
CBLN \cite{liu2021context} &CVPR 2021 &VGG &Glove &- &43.67 &24.44 & - & 88.39 &56.49\\
SV-VMR \cite{wu2021diving} &ICME 2021 & VGG &Glove &- &43.60 &24.78 & - &83.58 &50.22\\
MMN \cite{wang2022negative}&AAAI 2022 &VGG &DistilBERT &- &47.31 &27.28 & - & 83.74 &\underline{58.41} \\
UMT \cite{liu2022umt} &CVPR 2022 &VGG &Glove &- &48.31 &29.25 & - &\underline{88.79} &56.08\\
CDN \cite{wang2022cross} &TMM 2022 &VGG &Glove &- &45.24 &26.99 & - &81.18 &57.47\\
QD-DETR \cite{moon2023query} &CVPR 2023 &VGG &Glove &- &\underline{52.77} &\underline{31.31} &- &- &- \\
MomentDiff \cite{li2024momentdiff} & NeurIPS 2023 &VGG &Glove &- &51.94 &28.25 & - &- &-\\

\rowcolor{lightpink}
\textcolor{black}{TBSG-Net (\textbf{Ours})} &- &\textcolor{black}{VGG} &\textcolor{black}{GloVe} 
    & \textcolor{black}{{63.21} \tiny{($\uparrow$2.23\%)}}
    & \textcolor{black}{{53.03} \tiny{($\uparrow$0.49\%)}}
    & \textcolor{black}{{31.81} \tiny{($\uparrow$1.60\%)}}
    & \textcolor{black}{{96.30}}
    & \textcolor{black}{88.27 \tiny{($\downarrow$0.59\%)}}
    & \textcolor{black}{{59.46} \tiny{($\uparrow$1.80\%)}} \\

\rowcolor{lightpink}
TBSG-Net (\textbf{Ours}) &- &VGG &CLIP 
    & \textbf{66.53} \tiny{($\uparrow$7.60\%)}
    & \textbf{53.23} \tiny{($\uparrow$0.87\%)}
    & \textbf{32.24} \tiny{($\uparrow$2.97\%)}
    & \textbf{98.54}
    & \textbf{90.84} \tiny{($\uparrow$2.31\%)}
    & \textbf{61.16} \tiny{($\uparrow$4.71\%)} \\

\midrule

Moment-DETR \cite{lei2021detecting}&NIPS 2021 &CLIP &CLIP &- &55.65 &34.17 & - &- &-\\
UnLoc-L \cite{yan2023unloc}&ICCV 2023 &CLIP &CLIP &- &60.80 &38.40 & - & 88.20 &61.10\\
VDI \cite{luo2023towards} &CVPR 2023 &CLIP &CLIP &- &52.32 &31.37 & - &87.03 &62.30 \\
PaTF \cite{nguyen2024parallel} &ICMR 2024 & CLIP&CLIP &- & \underline{63.60}& \underline{40.80}&- &\underline{90.70} &\underline{65.30} \\

\rowcolor{lightpink}
TBSG-Net (\textbf{Ours}) &- &CLIP &CLIP & \textbf{74.83} & \textbf{63.77} \tiny{($\uparrow$0.27\%)} & \textbf{42.31} \tiny{($\uparrow$3.70\%)} & \textbf{98.85} & \textbf{92.75} \tiny{($\uparrow$2.26\%)} & \textbf{66.02} \tiny{($\uparrow$1.10\%)} \\

\midrule

\textcolor{black}{BM-DETR \cite{jung2025background}} & \textcolor{black}{WACV 2025} & \textcolor{black}{SF+CLIP} & \textcolor{black}{CLIP} & - & \textcolor{black}{59.48} & \textcolor{black}{38.33} & - & - & - \\
\textcolor{black}{OB-VMR \cite{li2026object}} & \textcolor{black}{AAAI 2026} & \textcolor{black}{SF+CLIP} & \textcolor{black}{CLIP} & \textcolor{black}{\underline{75.50}} & \textcolor{black}{\underline{65.10}} & \textcolor{black}{\underline{46.10}} & - & - & - \\
\rowcolor{lightpink}
\textcolor{black}{TBSG-Net (\textbf{Ours})} & - & \textcolor{black}{SF+CLIP} & \textcolor{black}{CLIP} 
    & \textcolor{black}{\textbf{77.90} \tiny{($\uparrow$3.18\%)}} 
    & \textcolor{black}{\textbf{67.83} \tiny{($\uparrow$4.20\%)}} 
    & \textcolor{black}{\textbf{46.52} \tiny{($\uparrow$0.91\%)}} 
    & \textcolor{black}{\textbf{99.10}} 
    & \textcolor{black}{\textbf{95.42}} 
    & \textcolor{black}{\textbf{68.91}} \\

\midrule

CBLN \cite{liu2021context} & CVPR 2021 & I3D & Glove & - & 61.13 & 38.22 & - & 90.33 & 61.69 \\
LPNet \cite{xiao2021natural} & arXiv 2021 & I3D & Glove & \underline{66.59} & 54.33 & 34.03 & - & - & - \\
SV-VMR~\cite{wu2021diving} & ICME 2021 & I3D & Glove & - & 55.55 & 32.75 & - & 89.01 & 56.18 \\
FVMR \cite{gao2021fast} & ICCV 2021 & I3D & Glove & - & 55.01 & 33.74 & - & 89.17 & 57.24 \\
UniMD \cite{zeng2025unimd} & ECCV 2024 & I3D & CLIP & - & 60.19 & 41.02 & - & \underline{91.61} & 65.86 \\
PaTF \cite{nguyen2024parallel} & ICMR 2024 & I3D & CLIP & - & \underline{64.00} & \underline{43.00} & - & 90.90 & \underline{66.50} \\

\rowcolor{lightpink}
TBSG-Net (\textbf{Ours}) & - & I3D & CLIP 
    & \textbf{76.60} \tiny{($\uparrow$15.03\%)} 
    & \textbf{65.46} \tiny{($\uparrow$2.28\%)} 
    & \textbf{44.85} \tiny{($\uparrow$4.30\%)} 
    & \textbf{98.92} 
    & \textbf{94.10} \tiny{($\uparrow$2.72\%)} 
    & \textbf{67.05} \tiny{($\uparrow$0.83\%)} \\

\bottomrule

\end{tabular}
\end{table*}
}

\subsection{Implementation Details}

\color{black}Our model is implemented by extending the UniMD~\cite{zeng2025unimd}
encoder–decoder framework for proposal-free VMR.
Specifically, we retain UniMD’s query-dependent classification and regression
heads, while augmenting the encoder with structured relational representations
derived from DSGs and the proposed TBSGs.

\noindent\textbf{Encoder.}
For each video, we extract coarse-grained 
frame-level features using pre-trained VGG, 
CLIP, I3D\textcolor{black}{, or SlowFast+CLIP} 
backbones, and textual queries are encoded using the CLIP text encoder 
in all settings\textcolor{black}{, except in the VGG 
group where a GloVe-based variant is additionally 
evaluated to enable controlled text-encoder comparisons.}
In parallel, 
fine-grained event-based clip-level DSGs are 
constructed offline as a preprocessing step 
using the STTran~\cite{cong2021spatial} model. 
These DSGs are temporally aggregated into 
clip-level TBSGs, which explicitly encode 
object--relation connectivity and temporal 
spans. The resulting TBSGs are processed by 
the TBSG Encoder module, consisting of a 
Transformer for global temporal modeling and 
a GCN for local relational reasoning. The 
encoded graph features are fused with 
coarse-grained frame-level features via 
weighted concatenation before being passed 
to the decoder.

\noindent\textbf{Decoder and Inference.}
Following UniMD, the decoder \textcolor{black}{takes $V_{\text{final}}$ and 
the textual feature $V_t$ as input, performs multimodal 
fusion, and predicts the target moment boundaries 
$(t^s, t^e)$ in a proposal-free manner}.
The classification head evaluates query-dependent temporal relevance, while the
regression head maps the query embedding to a convolutional kernel that interacts
with encoded video features to \textcolor{black}{predict the 
boundary offsets}.

\noindent\textbf{Training Details.}
The model is trained using the AdamW optimizer with a learning rate 
of $1 \times 10^{-4}$, for 100 epochs with batch size 4. 
The classification and regression losses are 
equally weighted 
($\lambda_{cls}=\lambda_{reg}=1.0$). 
All experiments are implemented in PyTorch 
2.4.0 with CUDA 12.1 and cuDNN 9.0.10 on 
a single NVIDIA RTX A5000 GPU (24 GB).

\subsection{Comparison with SOTA Methods}
To thoroughly evaluate TBSG-Net, we compare 
its retrieval performance against 
\textcolor{black}{19 unique competitive baselines evaluated 
across four backbone settings} on 
Charades-STA. For all baselines, we use the 
official text encoder provided by each method 
to ensure reproducibility and avoid 
re-implementation bias. Earlier VMR models~\cite{lu2019debug,zhang2019man,
liu2021context,wu2021diving,li2024momentdiff} were 
designed around GloVe-based RNN encoders, 
and substituting CLIP-text would require 
non-trivial architectural modifications 
(e.g., changes to their cross-modal matching 
modules), making controlled comparison 
difficult. Following the evaluation protocol 
commonly adopted in recent CLIP-based VMR 
works~\cite{lei2021detecting,yan2023unloc,luo2023towards,nguyen2024parallel,zeng2025unimd}, we compare our 
CLIP-based model against (1) older baselines 
using their official GloVe settings and 
(2) modern CLIP-based methods under the same 
multimodal embedding space. \textcolor{black}{To 
further enable a text-encoder-controlled 
comparison in the VGG group, we additionally 
report a GloVe variant of TBSG-Net 
(TBSG-Net, VGG+GloVe).} This ensures 
consistency with established practice in the 
VMR literature.

Table~\ref{tab:main-result} shows detailed 
results. \textcolor{black}{TBSG-Net consistently achieves SOTA 
results across all backbone settings.} 
\textcolor{black}{In the VGG group, TBSG-Net 
with GloVe text encoder already surpasses all 
GloVe-based baselines across all metrics, and 
the CLIP variant further improves upon this, 
confirming that both the DSG contribution and 
the stronger text encoder jointly drive 
performance gains.} \textcolor{black}{In the CLIP and I3D 
groups, TBSG-Net outperforms the previous 
best method PaTF~\cite{nguyen2024parallel} 
by consistent margins across all IoU 
thresholds, demonstrating the effectiveness 
of DSG-based relational reasoning over 
frame-level and coarse clip-level baselines.}

\textcolor{black}{\noindent\textbf{Comparison 
with Dual-Stream Methods (SF+CLIP).}
To further assess TBSG-Net under a stronger 
backbone setting, we evaluate an SF+CLIP 
dual-stream variant and compare against 
BM-DETR~\cite{jung2025background} and 
OB-VMR~\cite{li2026object}, both of which 
also employ SF+CLIP backbones. As shown in 
Table~\ref{tab:main-result}, TBSG-Net with 
SF+CLIP outperforms both methods across all 
R@1 and R@5 metrics. Notably, even the 
single-stream I3D variant of TBSG-Net 
surpasses both BM-DETR and OB-VMR on R@1 
IoU=0.3 and IoU=0.5, demonstrating that 
DSG-based relational reasoning provides 
consistent improvements that partially 
compensate for the absence of dedicated 
motion features. These results confirm that 
DSG-enhanced features provide a universal, 
backbone-agnostic mechanism for precise and 
robust VMR.}

\textcolor{black}{Beyond backbone generalisability, 
we further examine whether TBSG-Net is robust 
to temporal distribution shifts.} VMR models 
often suffer from temporal bias, with 
performance degrading when moment duration 
or positions shift~\cite{li2024momentdiff}. \textcolor{black}{We 
evaluate TBSG-Net's robustness under such 
distribution shifts using the anti-bias benchmarks 
Charades-STA-Len and Charades-STA-Mom, which 
introduce controlled shifts in moment duration 
and position, respectively.} As shown in 
Table~\ref{tab:antibias-result}, TBSG-Net 
achieves the highest scores and outperforms 
the strongest baseline, 
MomentDiff~\cite{li2024momentdiff}, by large 
margins across all IoU thresholds. \textcolor{black}{These gains are particularly pronounced at 
the strictest IoU threshold ($\mu = 0.7$), 
demonstrating TBSG-Net's ability to maintain 
precise boundary localization even when test 
moments lie outside the training distribution. Ablation results in 
Section~\ref{Ablation study} confirm that this 
robustness stems from the DSG generation and 
DSG-E modules, which model temporal 
relational structure at the clip level rather 
than relying on frame-level features.}

{\small
\begin{table*}[!t]
\caption{Performance comparisons between TBSG-Net and SOTA models on the anti-bias datasets Charades-STA-Len and Charades-STA-Mom, evaluated across moment duration and position. All experiments use VGG as feature backbone.\label{tab:antibias-result}}
\centering
\renewcommand\arraystretch{1}
\begin{tabular}{
  >{\centering\arraybackslash}p{2.3cm}|
  >{\centering\arraybackslash}p{2.1cm}
  >{\centering\arraybackslash}p{2.1cm}
  >{\centering\arraybackslash}p{2.1cm}|
  >{\centering\arraybackslash}p{2.1cm}
  >{\centering\arraybackslash}p{2.1cm}
  >{\centering\arraybackslash}p{2.1cm}
}
\toprule
\multirow{3}{*}{Method}  & \multicolumn{3}{c|}{Charades-STA-Len} & \multicolumn{3}{c}{Charades-STA-Mom}\\
\cline{2-7}
&\multicolumn{3}{c|}{R@1, IoU=$\mu$} &\multicolumn{3}{c}{R@1, IoU=$\mu$} \\
\cline{2-7}
& $\mu = 0.3$ & $\mu = 0.5$ &$\mu = 0.7$ & $\mu = 0.3$ & $\mu = 0.5$ &$\mu = 0.7$ \\
\midrule
2D-TAN \cite{zhang2020learning}  & 39.68 & 28.68 & 17.72 & 27.81 & 20.44 & 10.84 \\
MomentDETR \cite{lei2021detecting}  & 42.73 & 34.39 & 16.12 & 29.94 & 21.16 & 11.56 \\
MMN \cite{wang2022negative}  & 43.58 & 34.31 & 19.94 & 33.58 & 27.20 & 14.12 \\
MomentDiff \cite{li2024momentdiff}  & \underline{\textcolor{black}{51.25}}  & \underline{38.32} & \underline{23.38} & \underline{48.39} & \underline{33.59} & \underline{15.71} \\
\midrule
\rowcolor{lightpink}
TBSG-Net (\textbf{Ours})  & \textbf{57.50} ($\uparrow 10.05\%$) & \textbf{43.98} ($\uparrow 14.77\%$) & \textbf{25.99} ($\uparrow 11.16\%$) & \textbf{50.65} ($\uparrow 4.67\%$)& \textbf{42.67} ($\uparrow 27.03\%$)& \textbf{22.36} ($\uparrow 42.32\%$)\\
\bottomrule
\end{tabular}
\end{table*}
}

{\small
\begin{table*}[!t]
\centering
\caption{
Ablation study of model components on Charades-STA using I3D features. DSGs denotes the Dynamic Scene Graph generation module (Section~\ref{Dynamic Scene Graph Generation}); DSG-E denotes the Dynamic Scene Graph Embedding module, comprising Transformer and GCN submodules. (Section~\ref{Transformer variant module.}). The arrow \(\downarrow\) indicates performance degradation relative to the full model (final row). (Note: When both Transformer and GCN are \ding{55}, DSG-E is entirely removed 
  (no TBSG Constructor or Encoder). When either is \checkmark, the full 
  DSG-E module is present with the specified TBSG Encoder configuration.)
\label{tab:ablation_result_charades-sta}}
\renewcommand\arraystretch{1} 
\begin{tabular}{
  >{\centering\arraybackslash}p{0.8cm}
  >{\centering\arraybackslash}p{2.5cm}
  >{\centering\arraybackslash}p{2cm}|
  >{\centering\arraybackslash}p{2.4cm}
  >{\centering\arraybackslash}p{2.4cm}|
  >{\centering\arraybackslash}p{2.4cm}
  >{\centering\arraybackslash}p{2.4cm}
}
\toprule
\multicolumn{3}{c|}{Model Components} & \multicolumn{2}{c|}{R@1, IoU=$\mu$} &\multicolumn{2}{c}{R@5, IoU=$\mu$}\\
\midrule
\multirow{2}{*}{DSGs}& \multicolumn{2}{c|}{DSG-E} &\multirow{2}{*}{$\mu = 0.5$} &\multirow{2}{*}{$\mu = 0.7$} & \multirow{2}{*}{$\mu = 0.5$} &\multirow{2}{*}{$\mu = 0.7$} \\
\cline{2-3}
 &Transformer &GCN  & & & &\\
\midrule
\ding{55}  &\ding{55} &\ding{55} &54.21 ($\downarrow$ 17.19\%)  & 35.10 ($\downarrow$ 21.74\%) & 86.19 ($\downarrow$ 8.41\%) & 56.39 ($\downarrow$ 15.90\%) \\
\checkmark  &\ding{55} &\ding{55}&56.93 ($\downarrow$ 13.03\%) &39.21 ($\downarrow$ 12.58\%) & 88.19 ($\downarrow$ 6.28\%) & 57.20 ($\downarrow$ 14.69\%)\\
\checkmark  &\ding{55}  &\checkmark &58.02 ($\downarrow$ 11.37\%) & 40.10 ($\downarrow$ 10.59\%) & 89.92 ($\downarrow$ 4.44\%) & 60.09 ($\downarrow$ 10.38\%) \\
\checkmark  &\checkmark &\ding{55}   & \underline{62.32} ($\downarrow$ 4.80\%) & \underline{41.04} ($\downarrow$ 8.50\%) & \underline{92.44} ($\downarrow$ 1.76\%) & \underline{63.56} ($\downarrow$ 5.21\%) \\
\midrule
\rowcolor{lightpink} \checkmark  &\checkmark &\checkmark & \textbf{65.46} & \textbf{44.85} & \textbf{94.10} & \textbf{67.05}\\
\bottomrule
\end{tabular}
\end{table*}
}

{\small
\begin{table}[!t]
\color{black} 
\caption{Zero-shot transfer results on ActivityNet Captions validation set. All models are trained on Charades-STA only, with no fine-tuning on ActivityNet Captions.}
\label{tab:zeroshot}
\centering
\renewcommand\arraystretch{1.2}
\begin{tabular}{
  >{\centering\arraybackslash}p{2.0cm}|
  >{\centering\arraybackslash}p{1.0cm}
  >{\centering\arraybackslash}p{1.0cm}|
  >{\centering\arraybackslash}p{1.0cm}
  >{\centering\arraybackslash}p{1.0cm}
}
\toprule
\multirow{2}{*}{Method} & 
\multicolumn{2}{c|}{R@1, IoU=$\mu$} & 
\multicolumn{2}{c}{R@5, IoU=$\mu$} \\
\cline{2-5}
& $\mu$=0.5 & $\mu$=0.7 & $\mu$=0.5 & $\mu$=0.7 \\
\midrule
2D-TAN~\cite{zhang2020learning}     & 11.81 & 4.24 & 38.21 & 20.64 \\
MomentDiff~\cite{li2024momentdiff}  & 15.43 & 5.29 & 42.75 & 23.42 \\
UniMD~\cite{zeng2025unimd}          & 17.82 & 7.57 & 48.13 & 26.72 \\
\rowcolor{lightpink}
TBSG-Net            & \textbf{20.62} & \textbf{8.84} 
                                    & \textbf{49.41} & \textbf{27.25} \\
\bottomrule
\end{tabular}
\end{table}
}
\textcolor{black}{\subsection{Zero-Shot 
Generalization to ActivityNet Captions}\label{Zero-Shot 
Generalization to ActivityNet Captions}}

\textcolor{black}{To evaluate the 
cross-dataset generalisation of TBSG-Net, we 
conduct a zero-shot transfer experiment on 
ActivityNet Captions~\cite{krishna2017dense}. 
All models are trained exclusively on 
Charades-STA and evaluated on the ActivityNet 
Captions validation set without any 
fine-tuning or domain adaptation. Following 
the same offline preprocessing pipeline used 
for Charades-STA, we apply the pre-trained 
STTran to ActivityNet Captions videos to 
construct pseudo-label DSGs. As shown in 
Table~\ref{tab:zeroshot}, TBSG-Net 
consistently outperforms all baselines under 
this zero-shot setting. The performance gap 
between TBSG-Net and UniMD is smaller than 
in the in-domain setting, consistent with the 
reduced vocabulary overlap between Action 
Genome and ActivityNet scene content, which 
leads to sparser DSG inputs, yet the 
dual-stream design (Eq.~\ref{final_emb}) 
ensures competitive performance by falling 
back to the coarse-grained stream $V_g$ when 
DSG quality degrades. Nevertheless, the 
relational--temporal representations learned 
on Charades-STA transfer effectively, 
confirming the cross-dataset generalisability 
of the proposed framework.}

\color{black}\subsection{Ablation study}\label{Ablation study}

\subsubsection{Contributions of DSGs and DSG-E}

{\small
\begin{table*}[t]
\centering
\caption{Ablation study of model components on the Charades-STA-Len dataset using I3D features. \label{tab:ablation_result_charades-sta-len}}
\renewcommand\arraystretch{1} 
\begin{tabular}{
  >{\centering\arraybackslash}p{0.8cm}
  >{\centering\arraybackslash}p{2.5cm}
  >{\centering\arraybackslash}p{2cm}|
  >{\centering\arraybackslash}p{2.4cm}
  >{\centering\arraybackslash}p{2.4cm}|
  >{\centering\arraybackslash}p{2.4cm}
  >{\centering\arraybackslash}p{2.4cm}
}
\toprule
\multicolumn{3}{c|}{Model Components} & \multicolumn{2}{c|}{R@1, IoU=$\mu$} &\multicolumn{2}{c}{R@5, IoU=$\mu$}\\
\midrule
\multirow{2}{*}{DSGs}& \multicolumn{2}{c|}{DSG-E} &\multirow{2}{*}{$\mu = 0.5$} &\multirow{2}{*}{$\mu = 0.7$} & \multirow{2}{*}{$\mu = 0.5$} &\multirow{2}{*}{$\mu = 0.7$} \\
\cline{2-3}
 &Transformer &GCN  & & & &\\
\midrule
\ding{55}  &\ding{55} &\ding{55} & 39.17 ($\downarrow$ 25.77\%)  & 21.69 ($\downarrow$ 42.28\%) & 88.36 ($\downarrow$ 4.30\%) & 45.21 ($\downarrow$ 23.06\%) \\
\checkmark  &\ding{55} &\ding{55}&41.75 ($\downarrow$ 20.88\%) & 23.36 ($\downarrow$ 37.84\%) & 86.94 ($\downarrow$ 5.84\%) & 51.18 ($\downarrow$ 12.90\%)\\
\checkmark  &\ding{55}  &\checkmark & 43.54 ($\downarrow$ 17.49\%) & 24.92 ($\downarrow$ 33.69\%) & 87.82 ($\downarrow$ 4.88\%) & 50.91 ($\downarrow$ 13.36\%) \\
\checkmark  &\checkmark &\ding{55}   & \underline{46.32} ($\downarrow$ 12.22\%) & \underline{28.38} ($\downarrow$ 24.48\%) & \underline{88.65} ($\downarrow$ 3.99\%) & \underline{52.35} ($\downarrow$ 10.91\%) \\
\midrule
\rowcolor{lightpink} \checkmark  &\checkmark &\checkmark & \textbf{52.77} & \textbf{37.58} & \textbf{92.33} & \textbf{58.76}\\
\bottomrule
\end{tabular}
\end{table*}
}

{\small
\begin{table*}[t]
\centering
\caption{Ablation study of model components on the Charades-STA-Mom dataset using I3D features. \label{tab:ablation_result_charades-sta-mom}}
\renewcommand\arraystretch{1} 
\begin{tabular}{
  >{\centering\arraybackslash}p{0.8cm}
  >{\centering\arraybackslash}p{2.5cm}
  >{\centering\arraybackslash}p{2cm}|
  >{\centering\arraybackslash}p{2.4cm}
  >{\centering\arraybackslash}p{2.4cm}|
  >{\centering\arraybackslash}p{2.4cm}
  >{\centering\arraybackslash}p{2.4cm}
}
\toprule
\multicolumn{3}{c|}{Model Components} & \multicolumn{2}{c|}{R@1, IoU=$\mu$} &\multicolumn{2}{c}{R@5, IoU=$\mu$}\\
\midrule
\multirow{2}{*}{DSGs}& \multicolumn{2}{c|}{DSG-E} &\multirow{2}{*}{$\mu = 0.5$} &\multirow{2}{*}{$\mu = 0.7$} & \multirow{2}{*}{$\mu = 0.5$} &\multirow{2}{*}{$\mu = 0.7$} \\
\cline{2-3}
 &Transformer &GCN  & & & &\\
\midrule
\ding{55}  &\ding{55} &\ding{55}  & 36.05 ($\downarrow$ 32.30\%)  & 17.62 ($\downarrow$ 32.93\%) & 84.87 ($\downarrow$ 5.40\%)  & 43.44 ($\downarrow$ 16.77\%) \\
\checkmark  &\ding{55} &\ding{55}  & 40.97 ($\downarrow$ 23.06\%)  & 21.72 ($\downarrow$ 17.32\%)  & 86.08 ($\downarrow$ 4.05\%) & 45.60 ($\downarrow$ 12.63\%)  \\
\checkmark  &\ding{55}  &\checkmark & 41.43 ($\downarrow$ 22.20\%)  & 22.32 ($\downarrow$ 15.04\%) & \underline{87.90} ($\downarrow$ 2.02\%) & 46.10 ($\downarrow$ 11.67\%) \\
\checkmark  &\checkmark &\ding{55} & \underline{43.10} ($\downarrow$ 19.06\%) & \underline{23.18} ($\downarrow$ 11.76\%) & 87.24 ($\downarrow$ 2.70\%)  & \underline{47.38} ($\downarrow$ 9.22\%) \\
\midrule

\rowcolor{lightpink} 
\checkmark  &\checkmark &\checkmark& \textbf{53.25}  & \textbf{26.27} & \textbf{89.71} & \textbf{52.19} \\
\bottomrule
\end{tabular}
\end{table*}
}


We introduce the DSGs module to capture fine-grained event 
information at the clip level, complemented by the DSG-E module, which 
comprises \textbf{(1) a TBSG Constructor} that aggregates frame-level 
DSGs into clip-level temporal bipartite scene graphs, followed by 
\textbf{(2) a TBSG Encoder} that integrates a Transformer submodule 
for modeling global temporal dependencies and a GCN submodule for local 
structural reasoning. To evaluate each component's contribution, we 
conduct ablation studies on Charades-STA, with the main results in 
Table \ref{tab:ablation_result_charades-sta}.
\begin{itemize}
    \item \textbf{Without DSGs \& DSG-E}. Removing both modules results in the largest drop across all metrics, highlighting the importance of fine-grained event representation (DSGs) and effective spatio-temporal modeling (DSG-E).
    
    \item \textbf{Without DSG-E}. Removing the entire DSG-E 
    module (including both TBSG Constructor and TBSG Encoder) causes 
    the second-largest performance decline, indicating that the 
    construction and encoding of temporal bipartite scene graphs are 
    essential for leveraging DSG representations effectively.

    \item \textbf{Without Transformer in TBSG Encoder}. Removing only 
    the Transformer submodule while retaining the GCN significantly 
    reduces R@1, highlighting the Transformer's key role in modeling 
    long-range temporal dependencies across the bipartite graph structure.

    \item \textbf{Without GCN in TBSG Encoder}. Removing only the GCN 
    submodule while retaining the Transformer leads to a substantial 
    decrease, indicating that while the Transformer captures global 
    temporal relationships, the GCN is critical for refining local 
    relational structure and feature aggregation.

\end{itemize}

The ablation studies on Charades-STA-Len (Table \ref{tab:ablation_result_charades-sta-len}) 
and Charades-STA-Mom (Table \ref{tab:ablation_result_charades-sta-mom}) 
exhibit the same pattern as those on the original Charades-STA split. \textcolor{black}{The consistent gains across all three splits demonstrate 
that TBSG-Net does not overfit to dataset-specific temporal statistics, 
but instead captures generalizable relational–temporal structure.} Removing any component, whether DSGs, the 
entire DSG-E module, or individual submodules 
within it, consistently reduces performance and demonstrates the importance of each module for robust retrieval under changes in temporal distributions. The largest performance drop occurs when both DSGs and DSG-E are removed, highlighting their critical role in modeling varying moment durations and positions. The decreases caused by removing the Transformer and GCN confirm that both global self-attention and local graph convolution are necessary \textcolor{black}{components of the TBSG-Net’s minimal viable configuration, and are essential for mitigating length and position biases in VMR}.

\subsubsection{Impact of Visual Features}
{\small
\begin{table}[t]
\centering
\caption{\textcolor{black}{Ablation study on Charades-STA for 
different combinations of VGG, I3D, CLIP, \textcolor{black}{SF+CLIP,} 
and DSG modules. Percentage decreases are shown relative to the 
full \textcolor{black}{SF+CLIP}+DSGs model.}}
\label{tab:Different visual feature}
\renewcommand\arraystretch{1}
\begin{tabular}{
  >{\centering\arraybackslash}p{0.3cm}
  >{\centering\arraybackslash}p{0.3cm}
  >{\centering\arraybackslash}p{0.3cm}
  >{\centering\arraybackslash}p{0.9cm}|
  >{\centering\arraybackslash}p{0.5cm}|
  >{\centering\arraybackslash}p{1.7cm}|
  >{\centering\arraybackslash}p{1.7cm}
  }
\toprule
\multicolumn{5}{c|}{Model Components} & 
\multicolumn{2}{c}{R@1, IoU=$\mu$} \\
\midrule
\multicolumn{4}{c|}{Visual Feature} & 
\multirow{2}{*}{DSGs} &
\multirow{2}{*}{$\mu=0.5$} &
\multirow{2}{*}{$\mu=0.7$} \\
\cline{1-4}
VGG & I3D & CLIP & \textcolor{black}{SF+CLIP} & & & \\
\midrule
$\checkmark$ & \ding{55} & \ding{55} & \textcolor{black}{\ding{55}} & \ding{55} &
49.80 \tiny{($\downarrow$26.58\%)} & 30.10 \tiny{($\downarrow$35.30\%)} \\

$\checkmark$ & \ding{55} & \ding{55} & \textcolor{black}{\ding{55}} & $\checkmark$ &
53.23 \tiny{($\downarrow$21.52\%)} & 32.24 \tiny{($\downarrow$30.70\%)} \\
\midrule

\ding{55} & \ding{55} & $\checkmark$ & \textcolor{black}{\ding{55}} & \ding{55} &
61.52 \tiny{($\downarrow$9.30\%)} & 40.12 \tiny{($\downarrow$13.76\%)} \\

\ding{55} & \ding{55} & $\checkmark$ & \textcolor{black}{\ding{55}} & $\checkmark$ &
63.77 \tiny{($\downarrow$5.99\%)} & 42.31 \tiny{($\downarrow$9.05\%)} \\
\midrule

\ding{55} & $\checkmark$ & \ding{55} & \textcolor{black}{\ding{55}} & \ding{55} &
64.20 \tiny{($\downarrow$5.35\%)} & 43.00 \tiny{($\downarrow$7.57\%)} \\

\ding{55} & $\checkmark$ & \ding{55} & \textcolor{black}{\ding{55}} & $\checkmark$ &
\underline{65.46} \tiny{($\downarrow$3.49\%)} & 
\underline{44.85} \tiny{($\downarrow$3.59\%)} \\
\midrule

\textcolor{black}{\ding{55}} & \textcolor{black}{\ding{55}} & 
\textcolor{black}{\ding{55}} & \textcolor{black}{$\checkmark$} & 
\textcolor{black}{\ding{55}} &
\textcolor{black}{66.31 \tiny{($\downarrow$2.24\%)}} & 
\textcolor{black}{43.29 \tiny{($\downarrow$6.94\%)}} \\

\rowcolor{lightpink}
\textcolor{black}{\ding{55}} & \textcolor{black}{\ding{55}} & 
\textcolor{black}{\ding{55}} & \textcolor{black}{$\checkmark$} & 
\textcolor{black}{$\checkmark$} &
\textbf{\textcolor{black}{67.83}} & 
\textbf{\textcolor{black}{46.52}} \\
\bottomrule
\end{tabular}
\end{table}
}
\textcolor{black}{To understand how different levels of visual 
representation contribute to TBSG-Net, we evaluate four visual 
encoders: VGG, CLIP, I3D, \textcolor{black}{and SF+CLIP}. These 
encoders provide complementary perspectives on the video: VGG 
captures frame-level appearance cues, CLIP offers strong semantic 
alignment through its image--text pretraining, I3D explicitly models 
motion dynamics through 3D convolution, \textcolor{black}{and SF+CLIP 
combines SlowFast motion features with CLIP features for richer 
temporal representation.}}

\textcolor{black}{Table~\ref{tab:Different visual feature} summarizes 
the results. As expected, VGG-based variants yield the weakest 
performance due to VGG's lack of temporal modeling, but they provide 
a useful lower bound showing that TBSG-Net can still extract gains 
from DSGs even with appearance-only features. CLIP-based variants 
achieve higher accuracy by incorporating stronger semantic 
representations, which improve object and relation understanding. 
I3D-based variants further improve performance by introducing 
motion-sensitive temporal cues. \textcolor{black}{SF+CLIP-based 
variants achieve the strongest results by combining motion dynamics 
with semantically rich CLIP features, demonstrating that richer 
temporal visual features further amplify the benefits of DSG-based 
relational reasoning.}}

\textcolor{black}{Across all \textcolor{black}{four} backbones, 
integrating the DSGs module consistently improves performance, 
demonstrating the complementary role of fine-grained relational 
structure in enhancing visual features. \textcolor{black}{Among all 
configurations, SF+CLIP+DSGs achieves the strongest results, 
confirming the effectiveness of coupling rich temporal visual 
dynamics with structured event-level reasoning for accurate and 
robust video moment retrieval.}}

\textcolor{black}{We do not ablate the text encoder as replacing 
CLIP-text with a non-contrastive encoder (e.g., GloVe) would alter 
the cross-modal alignment objective itself, introducing a confounding 
variable rather than isolating the architectural contribution of 
TBSG-Net.}
{\small
\begin{table}[!t]
\centering
\caption{Effect of different relative placements of Transformer and GCN in Charades-STA: (a) GCN before Transformer, (b) Transformer before GCN, and (c) Operating in parallel.
\label{tab:relative position of Tran-GCN}}
\renewcommand\arraystretch{1} 
\begin{tabular} {
  >{\centering\arraybackslash}p{1.6cm}|
  >{\centering\arraybackslash}p{1.2cm}
  >{\centering\arraybackslash}p{1.2cm}|
  >{\centering\arraybackslash}p{1.2cm}
  >{\centering\arraybackslash}p{1.2cm}
}
\toprule
\multirow{2}{*}{Model Variants} &\multicolumn{2}{c|}{R@1, IoU=$\mu$} &\multicolumn{2}{c}{R@5, IoU=$\mu$} \\
\cline{2-5}
&$\mu$ = 0.5 &$\mu$ = 0.7 &$\mu$ = 0.5 &$\mu$ = 0.7 \\
\midrule
(a)& \underline{61.08} & \underline{39.52} & \underline{91.83} &\underline{64.08}\\
\rowcolor{lightpink}
(b)& \textbf{65.46} & \textbf{44.85} & \textbf{94.10} &\textbf{67.05} \\
(c)& 59.32 & 38.34 & 90.02 &63.12 \\

\bottomrule
\end{tabular}
\end{table}
}

\subsubsection{Impact of relative position of Transformer-GCN architecture}
We conduct an ablation study to examine how the stacking order (Fig. \ref{fig:relative_position}) of Transformer and GCN influences retrieval performance. The results in Table \ref{tab:relative position of Tran-GCN} reveal: \textbf{Model (a)} (GCN before Transformer) yields reasonable performance, indicating that aggregating local graph features before global attention is effective. \textbf{Model (b)} (Transformer before GCN) yields the best results, suggesting that capturing global dependencies first enhances feature representation, allowing the GCN to refine local interactions more effectively. \textbf{Model (c)} (Parallel Transformer-GCN) performs worst. This decline likely results from weakened feature interactions due to the lack of a sequential flow between global and local processing.

\begin{table}[t]
\centering
\caption{\textcolor{black}{Ablation study on Charades-STA evaluating different time-span encoding 
methods in the TBSG Constructor: (1) no explicit encoding, (2) binary time-span indicator, 
(3) randomized time spans, and (4) the duration-aware temporal weighting model.}}

\begin{tabular}{c|ccc|ccc}
\toprule
\multirow{2}{*}{\textcolor{black}{Variant}} & \multicolumn{3}{c|}{\textcolor{black}{R@1}} 
& \multicolumn{3}{c}{\textcolor{black}{R@5}} \\
\cline{2-7}
& \textcolor{black}{0.3} & \textcolor{black}{0.5} & \textcolor{black}{0.7} & \textcolor{black}{0.3} & \textcolor{black}{0.5} & \textcolor{black}{0.7} \\
\hline
\textbf{\textcolor{black}{(1)}} &
\textcolor{black}{72.60} & \textcolor{black}{61.50} & \textcolor{black}{38.05} &
\textcolor{black}{93.80} & \textcolor{black}{88.40} & \textcolor{black}{56.00} \\

\textbf{\textcolor{black}{(2)}} &
\textcolor{black}{74.10} & \textcolor{black}{62.80} & \textcolor{black}{40.12} &
\textcolor{black}{95.10} & \textcolor{black}{89.70} & \textcolor{black}{58.30} \\

\textbf{\textcolor{black}{(3)}} &
\textcolor{black}{70.30} & \textcolor{black}{59.20} & \textcolor{black}{35.50} &
\textcolor{black}{92.10} & \textcolor{black}{86.50} & \textcolor{black}{53.40} \\

\textbf{\textcolor{black}{(4)}} &
\textbf{\textcolor{black}{76.60}} & \textbf{\textcolor{black}{65.46}} & \textbf{\textcolor{black}{44.85}} &
\textbf{\textcolor{black}{98.92}} & \textbf{\textcolor{black}{94.10}} & \textbf{\textcolor{black}{67.05}} \\
\bottomrule
\end{tabular}
\label{tab:impact_decay}
\end{table}

\subsubsection{\textcolor{black}{Impact of the Duration-Aware Temporal Weighting Term $\mathbb{T}$}}

\textcolor{black}{The TBSG explicitly incorporates duration information for 
each object–relation interaction through the duration-aware weighting term 
\textcolor{black}{\(\mathbb{T}_{ij} = \exp\!\left(\lambda \cdot 
\frac{\mathcal{T}_{ij}}{T_{\max}}\right)\)}, where $\mathcal{T}_{ij}$ denotes the 
temporal span of the relationship and \textcolor{black}{$\lambda$ is a learnable scaling factor initialised at 0.1}. 
This term modulates the attention weights based on the persistence of each 
interaction, enabling the model to distinguish between transient and stable 
relational patterns. To assess its contribution, we evaluate four alternative 
time-span encoding strategies.}

\textcolor{black}{
\textbf{(1) No Time Span Encoding:}
The duration of each relation is removed, and only its occurrence interval 
is retained. Without explicit modeling of interaction persistence, the 
temporal signal becomes coarse and unable to distinguish between momentary 
and sustained events.}

\textcolor{black}{\textbf{(2) Binary Time Span Indicator:} 
We replace the continuous duration with a binary flag 
indicating whether a relation appears within the clip. 
This representation ignores the magnitude of the duration 
and provides minimal temporal information.}

\textcolor{black}{\textbf{(3) Randomized Time Spans:} 
To examine the importance of accurate temporal information, we retain the 
relational structure but replace true durations with randomized values. 
This disrupts the temporal coherence of the TBSG and directly tests the 
model's reliance on meaningful duration information.}

\textcolor{black}{\textbf{(4) Duration-Aware Weighting (Full Model):} 
The full TBSG-Net employs the exponential weighting formulation \(\mathbb{T}_{ij}\)
\textcolor{black}{defined above}, which smoothly emphasizes 
long-duration, stable interactions while placing relatively less weight 
on transient or short-lived relations. This provides a continuous, 
structure-aware temporal weighting mechanism aligned with fine-grained 
moment localization.
}

\textcolor{black}{
Table~\ref{tab:impact_decay} shows that the full model consistently 
delivers the best results. Removing duration information entirely (Variant (1)) 
leads to the most substantial performance degradation, while the Binary 
Indicator (Variant (2)) and Randomized Duration (Variant (3)) variants also 
exhibit noticeable drops. These findings demonstrate that accurate and 
continuous time-span encoding is essential for capturing the temporal 
evolution of relational structure within dynamic scene graphs and for 
enabling precise moment localization.
}

\begin{figure}[t!]
\centering
\includegraphics[width=2.5in]{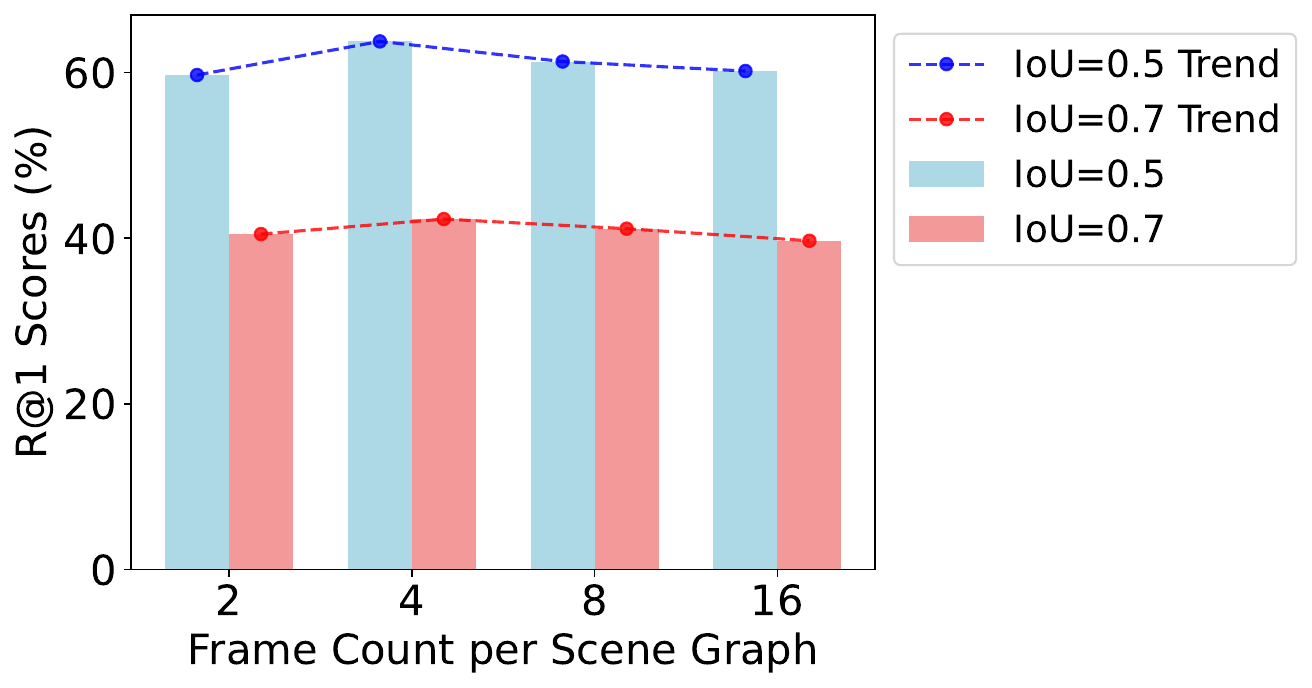}
\caption{Impact of the number of frames per dynamic scene graph on R@1 at IoU thresholds of 0.5 and 0.7 on Charades-STA.}

\label{Effect of Frame Count per Scene Graph on R1@k}
\end{figure}

\begin{figure}
\centering
\subfloat[Impact of the number of objects o]{\includegraphics[width=0.24\textwidth]{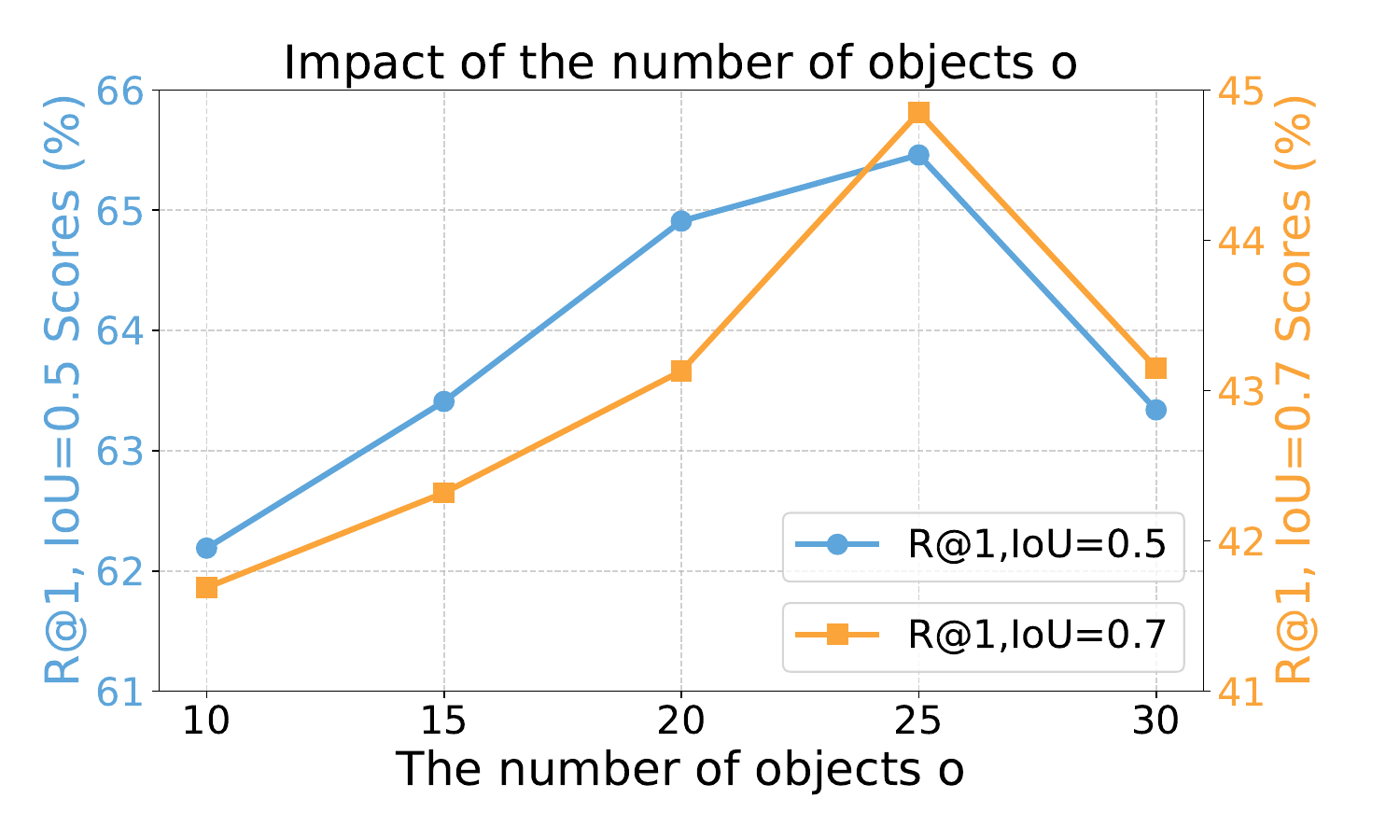}\label{fig:sub1}}
\hspace{0.02cm} 
\subfloat[Impact of the number of relationships r]{\includegraphics[width=0.24\textwidth]{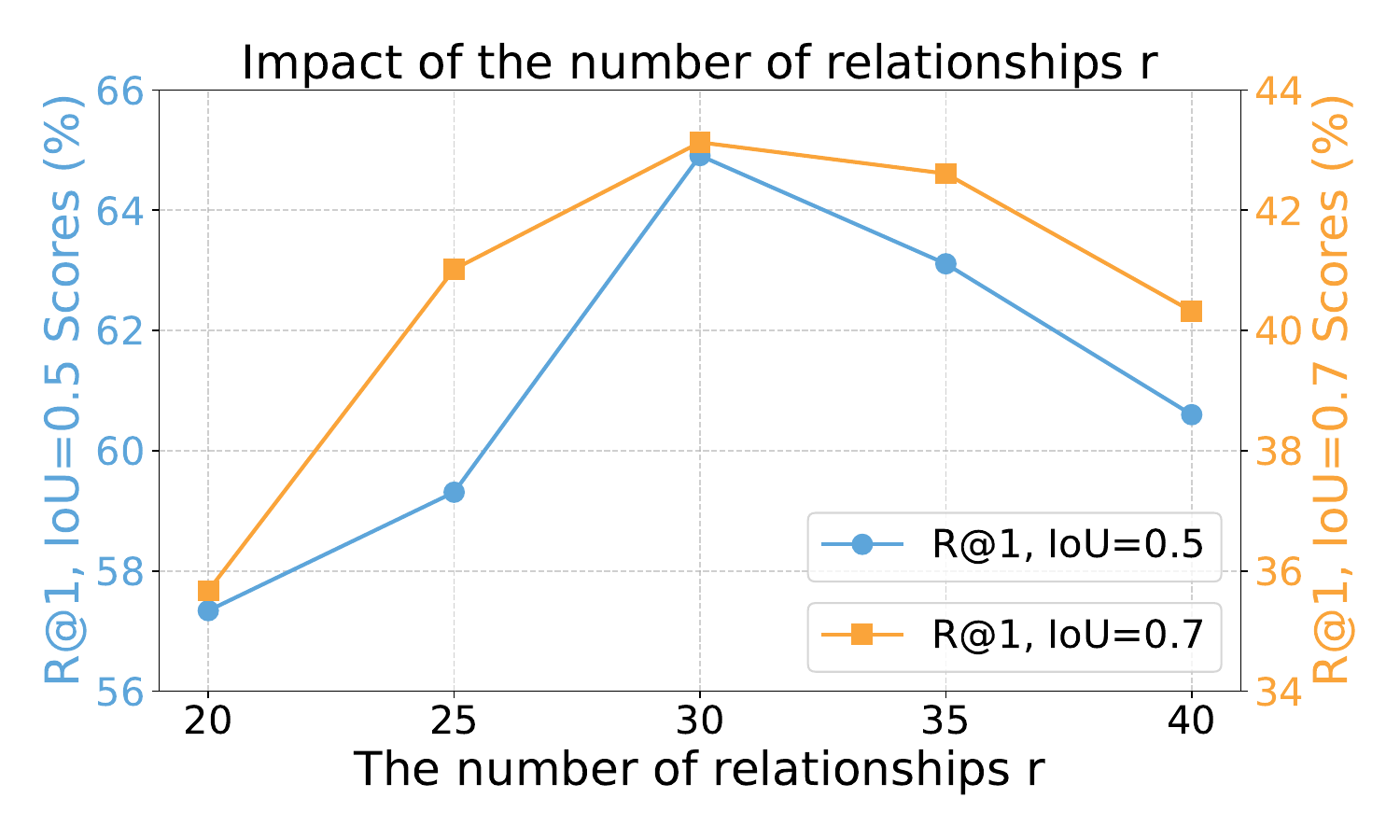}\label{fig:sub2}}

\caption{Effect of the maximum number of objects \(o\) and relationships \(r\) per DSG on retrieval performance (R@1 at IoU 0.5 and 0.7) on Charades-STA. \label{fig:o and r}}
\end{figure}

\begin{figure}[t!]
\centering

\subfloat[R@1, IoU=0.5]{\includegraphics[width=0.23\textwidth]{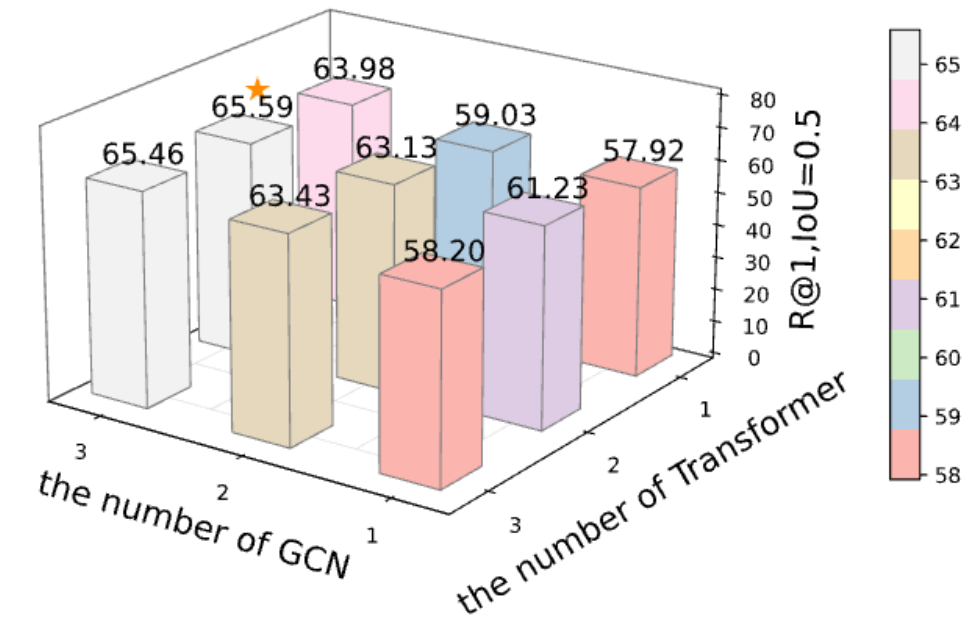}\label{fig:sub2}}
\subfloat[R@1, IoU=0.7]{\includegraphics[width=0.23\textwidth]{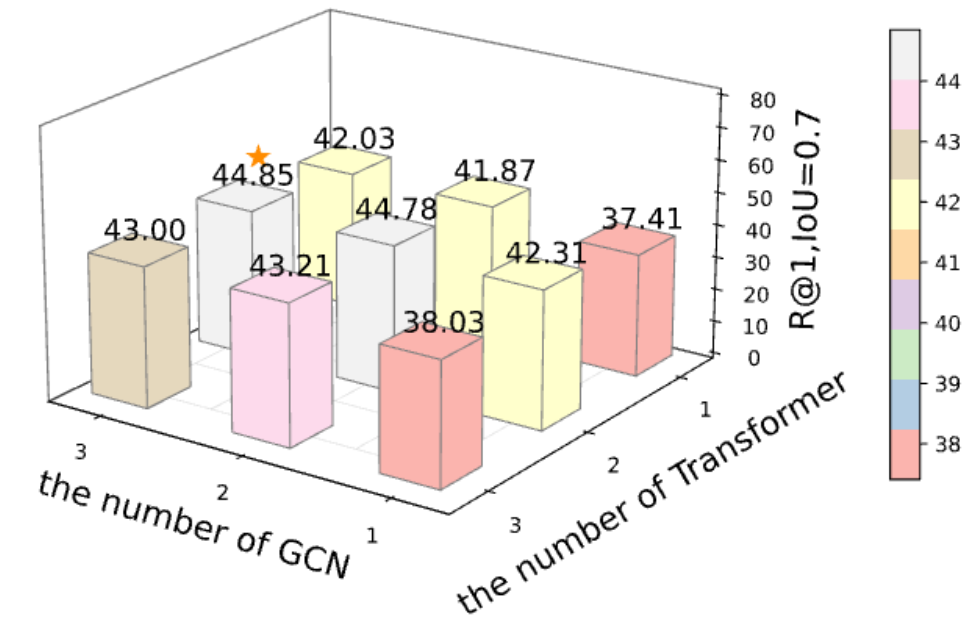}\label{fig:sub3}}

\caption{Effect of Transformer and GCN layer counts on Charades-STA. The x-axis represents the number of GCN layers, the y-axis represents the number of Transformer layers, and the z-axis indicates the R@1 score. The orange star $\star$ marks the configuration achieving the highest R@1 at IoU $\in \{0.5, 0.7\}$. \label{fig:GCN-Trans_number}}

\end{figure}

\subsection{Parameter study}\label{Parameter study}
\subsubsection{Impact of frame count per DSG}
DSGs generation module encodes objects and their relationships across multiple frames. To assess the impact of the frame count per DSG, we evaluate the performance of the model under varying frame numbers $F$ from 2 to 16, as shown in Fig. \ref{Effect of Frame Count per Scene Graph on R1@k}. Results show that performance improves up to 4 frames but declines thereafter, likely due to redundant or noisy spatio-temporal information hindering representation learning.

\subsubsection{Impact of the Number of Objects and Relationships}
Our model defines two key hyperparameters \(o\) (max objects per DSG) and \(r\) (max relationships per DSG). We vary these parameters and evaluate the retrieval performance, as shown in Fig.~\ref{fig:o and r} (a) and (b), performance improves as \(o\) and \(r\) increase, then plateaus around \(o=25\) and \(r=30\). Beyond these, adding more objects or relations yields negligible gains, suggesting the model captures the most salient scene information within these bounds, and more may add noise or redundancy.

\subsubsection{Impact of the number of Transformer and GCN}

\begin{table}[h]
\centering
\caption{\color{black}Impact of Upstream Confidence Thresholds on TBSG-Net Performance. The default setting is $P_{\text{obj}}{=}0.70$ and $P_{\text{rel}}{=}0.70$.}
\label{tab:sensitivity_analysis}
\resizebox{1.0\columnwidth}{!}{
\color{black}
\begin{tabular}{c|c|c}
\toprule
\textbf{Perturbation Variable (Threshold)} 
& \multicolumn{1}{c|}{\textbf{R@1, IoU=0.5}} 
& \multicolumn{1}{c}{\textbf{R@1, IoU=0.7}} \\
\midrule
\multicolumn{3}{c}{\textbf{A. Object Detection Confidence Sensitivity ($P_{\text{obj}}$)}} \\
\midrule
$P_{\text{obj}} \geq 0.60$          
& 65.10 ($\downarrow$0.55\%) 
& 44.50 ($\downarrow$0.78\%) \\
$P_{\text{obj}} \geq 0.65$          
& \textbf{65.49} ($\uparrow$0.04\%) 
& \textbf{45.10} ($\uparrow$0.56\%) \\
$P_{\text{obj}} \geq 0.70$          
& 65.46  
& 44.85  \\
$P_{\text{obj}} \geq 0.75$          
& 64.20 ($\downarrow$1.93\%) 
& 43.60 ($\downarrow$2.79\%) \\
$P_{\text{obj}} \geq 0.80$          
& 63.00 ($\downarrow$3.76\%) 
& 42.10 ($\downarrow$6.13\%) \\
\midrule
\multicolumn{3}{c}{\textbf{B. Relation Prediction Confidence Sensitivity ($P_{\text{rel}}$)}} \\
\midrule
$P_{\text{rel}} \geq 0.60$ 
& 65.00 ($\downarrow$0.70\%) 
& 44.40 ($\downarrow$1.00\%) \\
$P_{\text{rel}} \geq 0.65$          
& 65.10 ($\downarrow$0.55\%) 
& 44.60 ($\downarrow$0.56\%) \\
$P_{\text{rel}} \geq 0.70$ 
& \textbf{65.46}  
& \textbf{44.85}  \\
$P_{\text{rel}} \geq 0.75$          
& 64.60 ($\downarrow$1.31\%) 
& 44.10 ($\downarrow$1.67\%) \\
$P_{\text{rel}} \geq 0.80$          
& 63.90 ($\downarrow$2.38\%) 
& 43.50 ($\downarrow$3.01\%) \\
\bottomrule
\end{tabular}
}
\end{table}

\begin{table}[t]
\centering
\caption{\textcolor{black}{Sensitivity analysis of the loss weights $\lambda_{cls}$ and $\lambda_{reg}$ on Charades-STA. 
All metrics vary minimally across settings, indicating that TBSG-Net is robust to the choice of loss weights.}}
\resizebox{\linewidth}{!}{
\begin{tabular}{c|c|ccc|ccc}
\toprule
\multirow{2}{*}{\textcolor{black}{$\lambda_{cls}$}} & \multirow{2}{*}{\textcolor{black}{$\lambda_{reg}$}} 
& \multicolumn{3}{c|}{\textcolor{black}{R@1, IoU=$\mu$}} 
& \multicolumn{3}{c}{\textcolor{black}{R@5, IoU=$\mu$}} \\
\cline{3-8}
& & \textcolor{black}{$\mu$=0.3} & \textcolor{black}{$\mu$=0.5} & \textcolor{black}{$\mu$=0.7} & \textcolor{black}{$\mu$=0.3} & \textcolor{black}{$\mu$=0.5} & \textcolor{black}{$\mu$=0.7} \\
\midrule
\textcolor{black}{1.0} & \textcolor{black}{1.0} & 
\textcolor{black}{76.60} & 
\textcolor{black}{65.46} & 
\textcolor{black}{44.85} & 
\textcolor{black}{98.92} & 
\textcolor{black}{94.10} & 
\textcolor{black}{67.05} \\  

\textcolor{black}{0.5} & \textcolor{black}{1.0} & 
\textcolor{black}{76.28 (-0.32)} & 
\textcolor{black}{65.11 (-0.35)} & 
\textcolor{black}{44.62 (-0.23)} & 
\textcolor{black}{98.74 (-0.18)} & 
\textcolor{black}{93.85 (-0.25)} & 
\textcolor{black}{66.88 (-0.17)} \\

\textcolor{black}{1.5} & \textcolor{black}{1.0} & 
\textcolor{black}{76.55 (-0.05)} & 
\textcolor{black}{65.39 (-0.07)} & 
\textcolor{black}{44.78 (-0.07)} & 
\textcolor{black}{98.86 (-0.06)} & 
\textcolor{black}{94.03 (-0.07)} & 
\textcolor{black}{67.00 (-0.05)} \\

\textcolor{black}{2.0} & \textcolor{black}{1.0} & 
\textcolor{black}{76.41 (-0.19)} & 
\textcolor{black}{65.23 (-0.23)} & 
\textcolor{black}{44.71 (-0.14)} & 
\textcolor{black}{98.80 (-0.12)} & 
\textcolor{black}{93.97 (-0.13)} & 
\textcolor{black}{66.95 (-0.10)} \\

\textcolor{black}{1.0} & \textcolor{black}{0.5} & 
\textcolor{black}{76.22 (-0.38)} & 
\textcolor{black}{65.02 (-0.44)} & 
\textcolor{black}{44.59 (-0.26)} & 
\textcolor{black}{98.70 (-0.22)} & 
\textcolor{black}{93.81 (-0.29)} & 
\textcolor{black}{66.84 (-0.21)} \\

\textcolor{black}{1.0} & \textcolor{black}{1.5} & 
\textcolor{black}{76.51 (-0.09)} & 
\textcolor{black}{65.31 (-0.15)} & 
\textcolor{black}{44.80 (-0.05)} & 
\textcolor{black}{98.88 (-0.04)} & 
\textcolor{black}{94.00 (-0.10)} & 
\textcolor{black}{67.01 (-0.04)} \\

\textcolor{black}{1.0} & \textcolor{black}{2.0} & 
\textcolor{black}{76.44 (-0.16)} & 
\textcolor{black}{65.34 (-0.12)} & 
\textcolor{black}{44.73 (-0.12)} & 
\textcolor{black}{98.83 (-0.09)} & 
\textcolor{black}{94.01 (-0.09)} & 
\textcolor{black}{66.98 (-0.07)} \\
\bottomrule
\end{tabular}
}
\label{tab:lambda_sensitivity_full}
\end{table}

Fig. \ref{fig:GCN-Trans_number} shows how varying Transformer and GCN layers affects R@1 on Charades-STA. Key insights: (1) Increasing GCN depth (1$\rightarrow$3) boosts R@1 by better aggregating object–relation structures. (2) Transformer depth shows diminishing returns beyond 2 layers, suggesting deeper self-attention may introduce overfitting. Overall, 3 GCN + 2 Transformer layers are optimal for robust moment retrieval.  

\subsubsection{\textcolor{black}{Sensitivity to Upstream Noise}}
\textcolor{black}{To evaluate the robustness of TBSG-Net under imperfect 
upstream predictions, we conduct a sensitivity analysis by perturbing the 
confidence thresholds of the object detector ($P_{\text{obj}}$) and the 
relation predictor ($P_{\text{rel}}$) at inference time. Increasing these 
thresholds progressively removes low-confidence predictions, resulting in 
sparser DSGs and allowing us to assess how sensitive TBSG-Net is to reduced 
and incomplete upstream evidence.}

Table~\ref{tab:sensitivity_analysis} 
summarizes results, yielding three 
observations:

\textcolor{black}{\textbf{(1) Graceful degradation under stronger filtering.}
When $P_{\text{obj}}$ increases from 0.70 (default) to 0.80, 
\textcolor{black}{R@1 drops by at most 3.76\% and 6.13\% at 
IoU=0.5 and IoU=0.7 respectively.} Similar patterns are observed for relation thresholds. 
These results indicate that TBSG-Net is not brittle to systematic 
degradation of upstream detections; instead, its relational–temporal representation degrades smoothly 
rather than collapsing under upstream noise.}

\begin{table}[!ht]
\color{black} 
\small
\caption{Performance of TBSG-Net under relation node corruption at inference time on Charades-STA with I3D backbone. A proportion $\rho$ of relation node features $X_r$ are replaced with randomly sampled features from other relation classes. The last row (No DSG) reports the configuration without any DSG modules from Table~\ref{tab:ablation_result_charades-sta} as the theoretical lower bound.}
\label{tab:noise_injection}
\centering
\renewcommand\arraystretch{1.2}
\begin{tabular}{
  >{\centering\arraybackslash}p{2cm}|
  >{\centering\arraybackslash}p{1.1cm}
  >{\centering\arraybackslash}p{1.1cm}|
  >{\centering\arraybackslash}p{1.1cm}
  >{\centering\arraybackslash}p{1.1cm}
}
\toprule
\multirow{2}{*}{Corruption $\rho$} & 
\multicolumn{2}{c|}{R@1, IoU=$\mu$} & 
\multicolumn{2}{c}{R@5, IoU=$\mu$} \\
\cline{2-5}
& $\mu$=0.5 & $\mu$=0.7 & 
  $\mu$=0.5 & $\mu$=0.7 \\
\midrule
0\% (clean) & 65.46 & 44.85 & 94.10 & 67.05 \\
20\%        & 63.88 & 43.49 & 93.38 & 65.87 \\
50\%        & 59.12 & 40.32 & 91.14 & 62.34 \\
80\%        & 52.97 & 34.28 & 84.72 & 54.02 \\
\midrule
No DSG      & 54.21 & 35.10 & 86.19 & 56.39 \\
\bottomrule
\end{tabular}
\end{table}

\textbf{(2) Stable performance across a wide operating range.}
We observe the best performance at $P_{\text{obj}}=0.65$ and 
$P_{\text{rel}}=0.70$, which slightly outperforms the default configuration 
($P_{\text{obj}}=0.70$, $P_{\text{rel}}=0.70$) by only +0.04\%–0.56\% 
on R@1. These marginal improvements demonstrate that TBSG-Net does not 
require careful threshold tuning and maintains stable performance across 
a broad range of confidence thresholds.

\textbf{(3) Effectiveness of temporal aggregation.}
The stable performance across threshold variations empirically 
confirms that our temporal aggregation mechanism successfully 
filters frame-level noise. The TBSG Constructor's retention of 
only temporally consistent patterns (as described in 
Section~\ref{Dynamic Scene Graphs Embedding}) is validated by the minimal performance 
degradation observed under sparse upstream evidence.

These findings confirm that TBSG-Net 
is resilient to both upstream detection and 
relation-prediction noise, validating the 
reliability of our multi-stage pipeline.

\textcolor{black}{The above analysis focuses 
on sparse upstream outputs; to further 
evaluate robustness against \textit{incorrect} 
(rather than merely sparse) predictions, we 
conduct a complementary relation node 
corruption experiment below.}

\textcolor{black}{At inference time, a 
proportion $\rho$ of relation node features 
$X_r$ are randomly replaced with features 
sampled from other relation classes, directly 
simulating systematic misclassification by 
the upstream detector. The no-DSG 
configuration in Table~\ref{tab:ablation_result_charades-sta} 
serves as the theoretical performance lower 
bound, representing complete upstream 
detector failure where the DSG stream 
contributes nothing to retrieval. Results 
are reported in Table~\ref{tab:noise_injection}.}

\textcolor{black}{As shown in Table~\ref{tab:noise_injection}, 
performance degrades gracefully up to 
$\rho = 50\%$, remaining above the no-DSG 
lower bound (R@1 IoU=0.5: 59.12 vs.\ 54.21). 
At $\rho = 50\%$, performance still exceeds 
the no-DSG-E configuration (R@1 IoU=0.5: 
56.93), since the remaining 50\% of correct 
relation nodes continue to provide valid 
relational signals. At $\rho = 80\%$, 
performance falls marginally below the no-DSG 
lower bound, as heavily corrupted relational 
cues actively mislead the attention mechanism. 
Such extreme misclassification rates are 
unlikely under normal operating conditions, 
as corroborated by the stable performance 
in Table~\ref{tab:sensitivity_analysis}.}

\subsubsection{\textcolor{black}{Impact of the loss weights $\lambda_{cls}$ and $\lambda_{reg}$}}
\textcolor{black}{To verify whether TBSG-Net is sensitive to the choice of 
loss weights, we evaluate the influence of $\lambda_{cls}$ (classification loss) 
and $\lambda_{reg}$ (regression loss) by varying each weight within the range 
$\{0.5, 1.0, 1.5, 2.0\}$ while keeping all other settings unchanged. 
As shown in Table~\ref{tab:lambda_sensitivity_full}, the performance across 
all metrics exhibits minimal variation, with the largest fluctuation being 
only 0.44\%. The default setting $(\lambda_{cls}, \lambda_{reg}) 
= (1.0, 1.0)$ yields consistently strong results, and all tested configurations 
remain within a narrow performance band.}

\textcolor{black}{These observations confirm that TBSG-Net exhibits stable 
performance across different loss weight configurations, indicating that its 
effectiveness stems primarily from the relational–temporal modeling design 
rather than from careful tuning of loss-weight coefficients.}

\subsection{Efficiency and Scalability Analysis}
\begin{table}[t]
\centering
\caption{\color{black}TBSG Size Statistics on the Validation Set: Average (and Maximum) Unique Object Classes ($o$) and Unique Relation Groups ($r$) as a Function of Input Frame Count ($F$).}
\label{tab:tbsg_size_stats}
\resizebox{1\columnwidth}{!}{
\color{black}
\begin{tabular}{c|c|c|c|c}
\toprule
\textbf{Frames per DSG($F$)} & \textbf{Avg. Unique Classes ($o$)} & \textbf{Max. Unique Classes ($o$)} & \textbf{Avg. Unique Groups ($r$)} & \textbf{Max. Unique Groups ($r$)} \\
\midrule
$F = 2$  & 3.55 & 9  & 4.14  & 14 \\
$F = 4$  & 8.07 & 22 & 5.67  & 27 \\  
$F = 8$  & 8.70 & 24 & 8.16  & 29 \\  
$F = 16$ & 9.59 & 24 & 12.90 & 32 \\  
\bottomrule
\end{tabular}
}
\end{table}

To provide further insights into the computational properties of TBSG-Net, we 
analyze wall-clock training time, online inference latency breakdown, GPU 
memory footprint, and the size of the constructed TBSGs.

\textbf{Training and inference time.}
We instrumented the training loop using \texttt{torch.cuda.synchronize()} and
measured that each training epoch requires 9.21 minutes on a single NVIDIA RTX
A5000 GPU (batch size 4).

\textbf{Online inference latency breakdown.}
\textcolor{black}{Object detection, relation extraction, and DSG generation are 
performed offline and excluded from reported latency. The 
average online inference latency is 232.7 ms per video (batch 
size 1): TBSG construction and graph encoding account for 
180.3 ms, and the proposal-free localization stage requires 
52.4 ms.}

\textbf{GPU memory footprint.}
\textcolor{black}{Peak training memory is 4.85 GB (forward) and 3.00 GB 
(backward), remaining stable across epochs. Inference requires 
approximately 2.14 GB.}

\textbf{Graph size statistics.}
\textcolor{black}{Table~\ref{tab:tbsg_size_stats} summarizes TBSG sizes across 
the test set after temporal aggregation. As the temporal window 
grows from 2 to 16 frames, relation groups increase smoothly 
(4.14 to 12.90 on average) while object classes grow 
moderately (3.55 to 9.59) and remain bounded (maximum 24). 
The observed maxima ($o \leq 24$, $r \leq 32$) align with the 
hyperparameter analysis (Section~\ref{Parameter study}), 
where performance plateaus at $o = 25$ and $r = 30$, 
confirming that naturally occurring graph sizes fall within 
TBSG-Net's optimal operating range.}


\subsection{Case Study}
\begin{figure}[t!]
\centering
\includegraphics[width=3.5in]{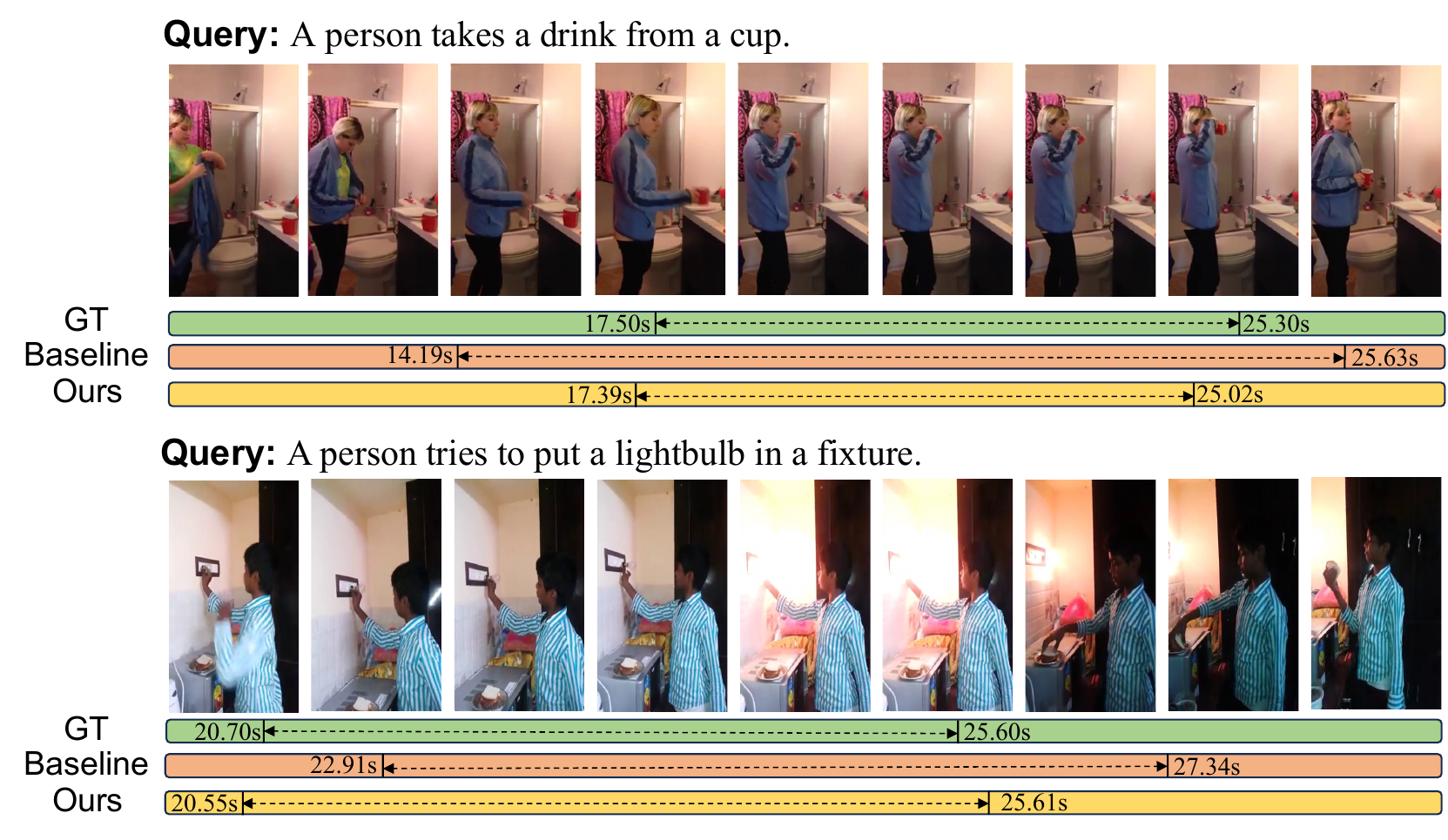}
\caption{Two Top-1 qualitative comparisons from Charades-STA. The three colored boxes indicate moment boundaries: green for ground truth (GT), orange for UniMD (baseline), and yellow for our model.
}
\label{Qualitative Analysis result}
\end{figure}
\textcolor{black}{Fig.~\ref{Qualitative Analysis result} presents two qualitative 
comparisons between TBSG-Net and UniMD on Charades-STA.}

In the first example, the query is \textit{A person takes a 
drink from a cup}. UniMD inaccurately retrieves moments that 
include \textit{trying to pick up the cup} and \textit{putting 
down the cup after drinking}. In contrast, our model with DSGs 
accurately identifies the relationship between \textit{the 
person} and \textit{the cup}, correctly excluding moments 
where the person is not touching the cup or has finished 
drinking, ensuring a precise match to the query.

In the second example, the query is \textit{A person tries to 
put a lightbulb in a fixture}. UniMD retrieves a broader 
temporal segment that extends beyond the point where the person 
appears finished with the attempt, and its localization of 
the target action's start time shows a delay. In 
contrast, our model accurately localizes the moment where the 
person's interaction with the \textit{lightbulb/fixture} is 
most relevant to the query.


\section{Conclusion and Future Work}
We have presented TBSG-Net, a novel DSG-based, 
proposal-free VMR framework. TBSG-Net leverages 
DSGs to extract graph-structured representations 
at the clip level, enabling the modeling of object 
interactions over time and capturing temporal 
dynamics. \textcolor{black}{Specifically, the TBSG 
Constructor aggregates frame-level DSGs into 
clip-level temporal bipartite scene graphs with 
explicit temporal span encoding, and the TBSG 
Encoder integrates a Transformer for global 
temporal modeling with a GCN for local relational 
reasoning.} Extensive experiments on Charades-STA, 
its two anti-bias variants, \textcolor{black}{and 
cross-dataset zero-shot transfer to ActivityNet 
Captions} demonstrate that TBSG-Net outperforms 
SOTA methods, particularly in handling complex, 
multi-step events with intricate temporal 
dependencies.

Several promising directions remain for future 
work. \textcolor{black}{At the data level,} we plan 
to extend TBSG-Net to a broader range of datasets 
through pseudo-labelling or semi-supervised 
annotation pipelines, reducing the current 
dependency on Action Genome vocabulary coverage. 
\textcolor{black}{At the architectural level, we 
aim to introduce an explicit directed adjacency 
formulation separating $A_{or}^{\text{sub}}$ and 
$A_{or}^{\text{obj}}$ into subject-role and 
object-role matrices to strengthen asymmetric 
relational reasoning on multi-agent datasets. 
Furthermore, extending the duration scaling factor 
$\lambda$ from a single shared scalar to a 
relation-type-specific or head-specific parameter 
within the multi-head attention framework would 
allow finer-grained adaptation to datasets with 
diverse event densities or dramatically different 
temporal scale distributions.} Finally, we aim to 
enhance scene graph representations by 
incorporating additional object attributes, such 
as color and state, to capture richer contextual 
information and improve retrieval precision.

\bibliographystyle{IEEEtran}
\bibliography{reference}
\newpage

\vfill

\end{document}